\documentclass{article}

\usepackage{microtype}
\usepackage{graphicx}
\usepackage{subfigure}
\usepackage{booktabs} 
\usepackage{lmodern}

\usepackage{hyperref}

\usepackage[table]{xcolor}
\usepackage[accepted]{icml2025}

\usepackage{amsmath}
\usepackage{amssymb}
\usepackage{mathtools}
\usepackage{amsthm}
\usepackage{multirow}

\usepackage{color}
\definecolor{mygreen}{RGB}{146, 199, 113} 
\usepackage{graphicx}
\usepackage{adjustbox}
\usepackage{array}
\usepackage{booktabs}
\usepackage{dsfont}
\usepackage{makecell}
\usepackage[capitalize,noabbrev]{cleveref}

\theoremstyle{plain}

\theoremstyle{definition}

\theoremstyle{remark}

\usepackage[textsize=tiny]{todonotes}

\icmltitlerunning{KIND: Knowledge Integration and Diversion for Training Decomposable Models}

\begin{document}

\twocolumn[
\icmltitle{KIND: Knowledge Integration and Diversion for Training Decomposable Models}



\icmlsetsymbol{equal}{*}

\begin{icmlauthorlist}
\icmlauthor{Yucheng Xie}{add1,add2}
\icmlauthor{Fu Feng}{add1,add2}
\icmlauthor{Ruixiao Shi}{add1,add2}
\icmlauthor{Jing Wang}{add1,add2}
\icmlauthor{Yong Rui}{add3}
\icmlauthor{Xin Geng}{add1,add2}
\end{icmlauthorlist}

\icmlaffiliation{add1}{School of Computer Science and Engineering, Southeast University, Nanjing, China}
\icmlaffiliation{add2}{Key Laboratory of New Generation Artificial Intelligence Technology and Its Interdisciplinary Applications (Southeast University), Ministry of Education, China}
\icmlaffiliation{add3}{Lenovo Research}

\icmlcorrespondingauthor{Jing Wang}{wangjing91@seu.edu.cn}
\icmlcorrespondingauthor{Xin Geng}{xgeng@seu.edu.cn}

\icmlkeywords{Machine Learning, ICML}

\vskip 0.3in
]



\printAffiliationsAndNotice{}  

\begin{abstract}
Pre-trained models have become the preferred backbone due to the increasing complexity of model parameters. However, traditional pre-trained models often face deployment challenges due to their fixed sizes, and are prone to negative transfer when discrepancies arise between training tasks and target tasks.
To address this, we propose \textbf{KIND}, a novel pre-training method designed to construct decomposable models.
KIND integrates knowledge by incorporating Singular Value Decomposition (SVD) as a structural constraint, with each basic component represented as a combination of a column vector, singular value, and row vector from $U$, $\Sigma$, and $V^\top$ matrices.
These components are categorized into \textbf{learngenes} for encapsulating class-agnostic knowledge and \textbf{tailors} for capturing class-specific knowledge, with knowledge diversion facilitated by a class gate mechanism during training.
Extensive experiments demonstrate that models pre-trained with KIND can be decomposed into learngenes and tailors, which can be adaptively recombined for diverse resource-constrained deployments. 
Moreover, for tasks with large domain shifts, transferring only learngenes with task-agnostic knowledge, when combined with randomly initialized tailors, effectively mitigates domain shifts.
Code will be made available at~\href{https://github.com/Te4P0t/KIND}{https://github.com/Te4P0t/KIND}.

\end{abstract}

\section{Introduction}
The increasing size of models has significantly increased computational costs, making pre-trained models a cornerstone of modern machine learning~\cite{qiu2020pre, han2021pre, feng2024redefining}. 
These pre-trained models have proven highly effective, especially when combined with parameter-efficient fine-tuning (PEFT) techniques such as LoRA~\cite{hulora, hayoulora} and its variants~\cite{zhang2023increlora, valipour2023dylora, liudora}. 
However, traditional pre-training approaches primarily focus on optimizing performance for specific training datasets, often neglecting their transferability to downstream tasks and adaptability to diverse deployment scenarios.

\begin{figure}[tb]
  \centering
  \includegraphics[width=\linewidth]{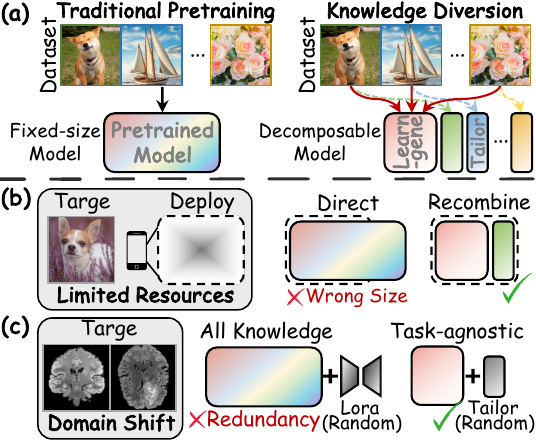}
  \vspace{-0.2in}
  \caption{(a)~Traditional pre-training prioritizes maximizing performance on training datasets, often producing fixed-size models and making them prone to negative transfer. In contrast, KIND redefines the training objective to pre-train models that are both structure- and knowledge-decomposable.
  (b)~Consequently, KIND enables pre-trained models to be adaptively restructured, facilitating deployment in diverse resource-constrained scenarios.
  (c)~Additionally, the task-agnostic knowledge encapsulated in learngenes can effectively mitigate domain shifts.}
  \label{fig:intro}
  \vspace{-0.1in}
\end{figure}

As a result, pre-trained models typically have a fixed, large size, designed to encapsulate as much knowledge as possible from the training data.
This design, however, presents significant challenges for practical deployment, which is often constrained by factors like memory usage, processing power, and response time~\cite{zhang2022minivit}.
More importantly, when downstream tasks differ significantly from the pre-training datasets, the transferred knowledge can become redundant~\cite{feng2024transferring}, biased~\cite{ren2024art}, or even harmful~\cite{wang2019characterizing, rosenstein2005transfer}.
These limitations underscore that traditional pre-trained models may not always serve as optimal backbones, as illustrated in Figure~\ref{fig:intro}.
This raises a critical question: \textit{Can we rethink the pre-training process to develop \textbf{decomposable pre-trained models} that can be adaptively adjusted to meet the specific requirements of downstream tasks and deployment scenarios?}

Recently, a novel knowledge transfer framework called \textit{Learngene} has been introduced~\cite{wang2023learngene}. 
Unlike traditional transfer learning methods, \textit{Learngene} encapsulates task-agnostic knowledge into modular network fragments~\cite{feng2023genes} known as learngenes, to enhance the efficiency of knowledge transfer and improve network adaptability.
Building upon the \textit{Learngene} framework, we propose KIND, a novel pre-training method that performs \textbf{K}nowledge \textbf{IN}tegration and \textbf{D}iversion during the pre-training process.
KIND is designed to construct flexible and decomposable pre-trained models, facilitating adaptive transformations to address the diverse requirements of downstream tasks and deployment scenarios.

KIND decomposes the weight matrix into \textit{basic components} for knowledge integration, then associates class-specific and class-agnostic knowledge with distinct components to facilitate knowledge diversion.
For this decomposition, KIND employs Singular Value Decomposition (SVD), representing each basic component as a combination of a column vector, singular value, and row vector derived from the $U$, $\Sigma$, and $V^\top$ matrices. 
These basic components are categorized into two types: \textbf{learngenes}, which encapsulate class-agnostic knowledge, and \textbf{tailors}, which capture class-specific knowledge.
Instead of directly applying SVD to pre-trained model weights~\cite{han2023svdiff, zhang2024spectral, robb2020few}, KIND incorporates SVD as a structural constraint during pre-training and trains the basic components rather than the full weight matrices. 
Such indirect training enables more explicit control over each class-specific component, guided by a class gate mechanism, thereby facilitating effective knowledge diversion.

We conduct experiments on class-conditional image generation tasks to better demonstrate knowledge transfer, using Diffusion Transformers (DiTs)~\cite{peebles2023scalable} as the backbone for diffusion models.
We pre-train DiT-B and DiT-L with KIND on ImageNet-1K, resulting in decomposable models that can be effectively divided into learngenes and tailors. 
Extensive experiments evaluate KIND across three scenarios. 
1) \textbf{General Tasks}: Models pre-trained with KIND perform on par with traditional pre-trained models (often outperforming them) without additional computational costs.
2) \textbf{Resource-constrained Scenarios}: KIND facilitates flexible combinations of learngenes and tailors to meet storage and computational limits, maintaining performance without sacrificing performance.
3) \textbf{Tasks with Large Domain Shifts}: KIND transfers learngenes only, combined with randomly initialized tailors, enabling efficient adaptation via class-agnostic knowledge.

Our main contributions are as follows: 
1) We redefine the pre-training objective by shifting the focus from solely maximizing model performance to diverting knowledge into class-agnostic knowledge and class-specific components, facilitating the construction of a more flexible and decomposable backbone adaptable to various scenarios.
2) We propose KIND, a novel pre-training method that integrates and diverts knowledge, marking the first application of learngenes to image generation tasks.
3) We establish a new benchmark for evaluating transfer efficiency and flexibility in diffusion models. Extensive experiments demonstrate that KIND achieves state-of-the-art performance while providing flexible storage and computational efficiency.

\section{Related Work}
\subsection{Initialization and Training of Variable-sized Models}
Practical deployments often encounter constraints related to memory usage, processing power, and response time, necessitating models of variable sizes~\cite{zhang2022minivit}. 
However, traditional pre-trained models are typically fixed in size, requiring \textit{\textbf{retraining}} when a suitable model size is unavailable~\cite{qiu2020pre, han2021pre}. 
While traditional model compression techniques, such as knowledge distillation~\cite{Gou2021Knowledge, muralidharan2024compact} and model pruning~\cite{zhang2024laptop, castells2024ld}, can generate models of variable sizes, they involve \textit{\textbf{repeated operations}} for each model size, resulting in significant inefficiencies in both time and resource consumption.

The \textit{Learngene} framework, inspired by the transfer of genetic information in nature~\cite{feng2023genes}, encapsulates common knowledge into modular network fragments, termed ``learngenes'', and employs them to initialize variable-sized models~\cite{wang2023learngene}. Notably, the process of condensing knowledge from pre-trained models into learngenes incurs a \textit{\textbf{one-time cost}}, eliminating the need for further training during model initialization.
Current learngene-based methods, either direct transfer selected layers from pre-trained models~\cite{wang2022learngene, wang2023learngene}, or apply predefined rules (e.g., Kronecker products) to distill knowledge knowledge into learngenes~\cite{xia2024transformer, feng2024wave}.
However, these approaches neglect the alignment between model components and their corresponding knowledge, limiting their efficiency and adaptability.

In contrast, KIND enhances such alignment through knowledge diversion during pre-training, constructing a decomposable model that enables more flexible and efficient initialization across varying model sizes.

\subsection{Parameter Efficient Fine-Tuning (PEFT)}
The increasing complexity of model parameters has made fine-tuning all parameters of pre-trained models resource-intensive and time-consuming~\cite{touvron2021training, achiam2023gpt}.
To address this,  PEFT techniques are developed to adapt large pre-trained models to new tasks by fine-tuning only a small set of parameters~\cite{hulora, houlsby2019parameter, hu2023llm, chen2022adaptformer}. 
Recent approaches apply SVD to pre-trained weight matrices, fine-tuning models by adjusting singular values, a process known as spectral shift~\cite{han2023svdiff, robb2020few, sun2022singular}, or by fine-tuning singular vectors~\cite{zhang2024spectrum, zhang2024spectral}.
However, existing PEFT methods rely on models pre-trained with traditional objectives and do not fully consider their adaptability as universal backbones across diverse tasks.

In contrast, KIND decomposes pre-trained models into learngenes and tailors through knowledge diversion. 
The class-agnostic knowledge encapsulated in learngenes significantly enhances transfer adaptability, particularly for tasks with large domain shifts compared to the training tasks.

\section{Methods}
\subsection{Preliminary}
\subsubsection{Latent Diffusion Models}
Latent diffusion models transfer the diffusion process from the high-resolution pixel space to the latent space by employing an autoencoder $\mathcal{E}$, which encodes an image $x$ into a latent code $z = \mathcal{E}(x)$. A diffusion model is then trained to generate the corresponding latent code in a denoising process, minimizing the following objective:
\begin{equation}
    \mathcal{L}=\mathbb{E}_{z,c,\varepsilon,t}[||\varepsilon-\varepsilon_{\theta}(z_t|c, t)||^{2}_{2}]
\label{eq:loss}
\end{equation}
Here, $\varepsilon_{\theta}$ is a noise prediction network that predicts the noise $\varepsilon$ added to $z_t$ at timestep $t$, conditioned on $c$.

\subsubsection{Diffusion Transformers (DiTs)}
DiT is a transformer-based architecture for noise prediction, replacing the traditional UNet. 
Given an image $x \in \mathbb{R}^{H_1 \times H_2 \times C}$ and its latent code $z \in \mathbb{R}^{h_1 \times h_2 \times c}$ encoded by $\mathcal{E}$, DiT divides the latent code $z$ into $T$ patches, which are then mapped to $D$-dimensional patch embeddings, with added position embeddings. 

The structure of DiTs resembles that of Vision Transformers (ViTs), which comprises $L$ stacked layers, each containing a Multi-Head Self-Attention (MSA) mechanism and a Pointwise Feedforward (PFF) layer. 
In each layer, a self-attention head $A_i$ performs self-attention using a query $Q$, key $K$, and value $V \in \mathbb{R}^{T \times D}$, with parameter matrices $W_q^i$, $W_k^i$, and $W_v^i \in \mathbb{R}^{D \times d}$:
\begin{equation}
    A_i = \text{softmax}(\frac{Q_i K_i^\top}{\sqrt{d}})V_i\,,\; A_i \in \mathbb{R}^{T \times d}
\end{equation}
MSA mechanism combines $h$ self-attention heads $A$ and projects the concatenated outputs using a weight matrix $W_o$:
\begin{equation}
    \text{MSA} = \text{concat}(A_1, A_2, ..., A_{h}) W_o\,,\; W_o \in \mathbb{R}^{hd \times D}
\label{equ:msa}
\end{equation}
In the implementation of MSA, the matrices $W_q^i$, $W_k^i$, and $W_v^i \in \mathbb{R}^{D \times d}$ for $h$ attention heads are combined into three parameter matrices $W_q$, $W_k$, and $W_v \in \mathbb{R}^{D \times hd}$.

PFF layer comprises two linear transformations $W_{in} \in \mathbb{R}^{D \times D'}$ and $W_{out} \in \mathbb{R}^{D' \times D}$ with a GELU~\cite{hendrycks2016gaussian} activation function:
\begin{equation}
    \text{PFF}(x) = \text{GELU}(xW_{in} + b_1)W_{out} + b_2
\label{equ:mlp}
\end{equation}
where $b_1$ and $b_2$ are the biases for the linear transformations, and $D'$ denotes the hidden layer dimensions.

\subsection{Knowledge Integration in Weight Matrices}
\begin{figure*}[tb]
  \centering
  \includegraphics[width=\linewidth]{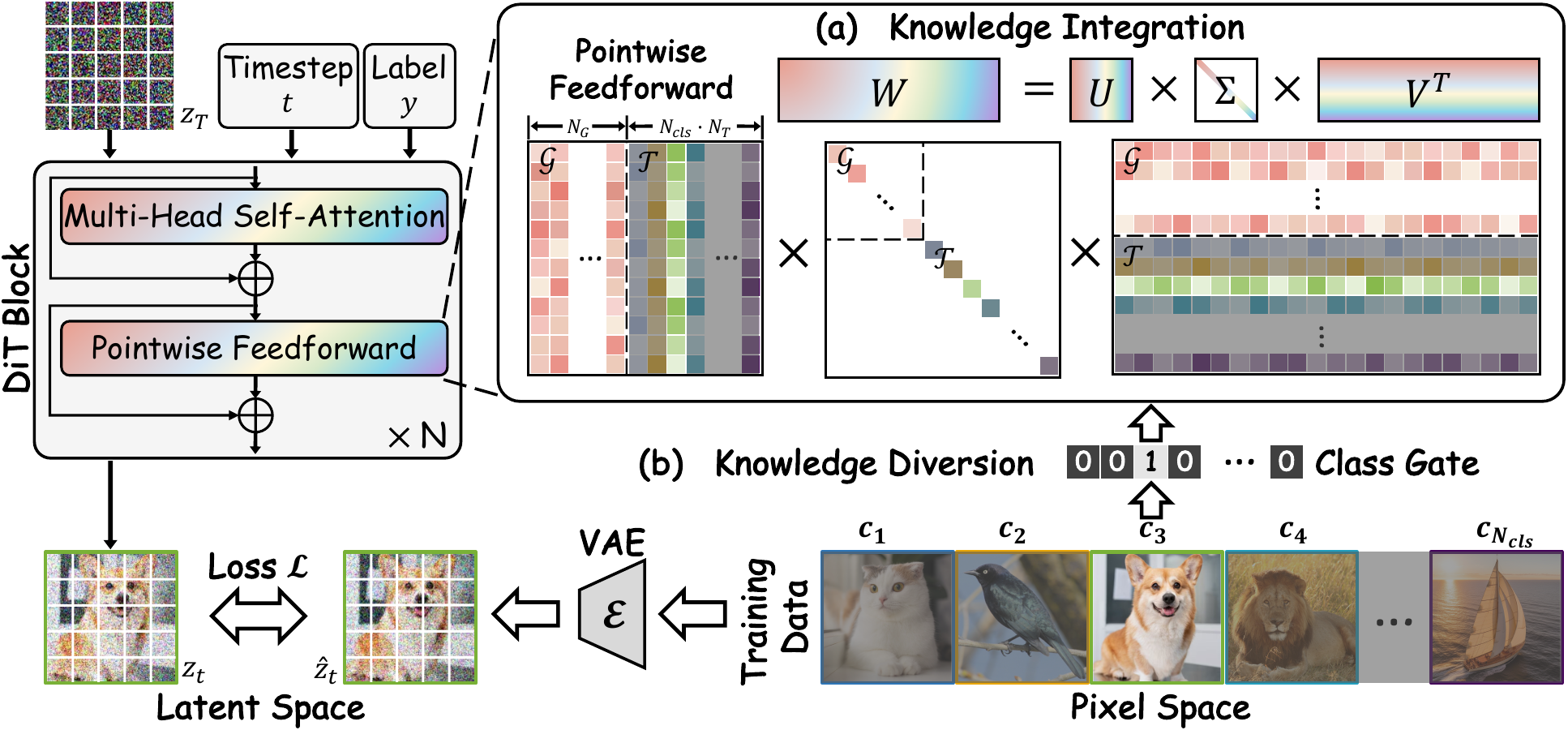}
  \vspace{-0.3in}
  \caption{(a) For each weight matrix in DiTs, we integrate it into the product of matrices $U$, $\Sigma$ and $V^\top$, formally inspired by SVD. 
  The components of these matrices are then explicitly partitioned into the learngenes and tailors, which encapsulate class-agnostic and class-specific knowledge, respectively.
  (b) Knowledge is diverted through a class gate ensuring each training image updates only the learngenes and their corresponding class-related tailors, so that the class-agnostic knowledge can be condensed into the learngenes, while knowledge specific to each class is diverted into corresponding tailors.}
  \label{fig:structure}
  \vspace{-0.1in}
\end{figure*}

FSGAN~\cite{robb2020few} directly applies SVD to pre-trained model parameters and fine-tunes the singular values for adaptation, achieving success in image segmentation~\cite{sun2022singular} and generation~\cite{han2023svdiff}
This shows that SVD can create a compact parameter space, facilitating efficient fine-tuning of pre-trained models. 

However, directly applying SVD to pre-trained parameter matrices decomposes them based on fixed orthogonalization rules, leading to poor interpretability and making it challenging to determine whether the knowledge in each basic component is class-specific.
This limits the model's decomposability, risking the loss of valuable knowledge.

To address this, we integrate knowledge by reconstructing weight matrices using the SVD-derived components $U$, $\Sigma$, and $V$, where each basic component is a combination of a column vector, singular value and row vector from $U$, $\Sigma$, and $V^\top$. We then explicitly associate each basic component with a specific type of knowledge (either class-specific or class-agnostic), which is achieved through a class gate mechanism to divert knowledge (Section~\ref{sec:know div}).

For the DiT architecture, the main weight matrices across the $L$-layers are $\theta = \{W_{q}^{(1\thicksim L)},$ $W_{k}^{(1\thicksim L)},$ $W_{v}^{(1\thicksim L)},$ $W_{o}^{(1\thicksim L)},$ $W_{in}^{(1\thicksim L)},$ $W_{out}^{(1\thicksim L)}\}$\footnote{$W_{q}^{(1\thicksim L)}$ denotes the set $\{W_{q}^{(1)}, W_{q}^{(2)}, \dots, W_{q}^{(L)}\}$. Similar notations throughout the paper follow this convention.}. 
Let $W_{\star}^{(l)}$ represent any weight matrix in layer $l$, where $\star \in \mathcal{S}$ and $\mathcal{S} = \{q,$ $k,$ $v,$ $o,$ $in,$ $out\}$ denotes the set of subscripts. The matrices $U_{\star}^{(l)}$, $\Sigma_{\star}^{(l)}$, $V_{\star}^{(l)}$ are the corresponding components that constitute $W_{\star}^{(l)}$, which is calculated as:
\begin{equation}
\begin{aligned}
    W_{\star}^{(l)} &= U_{\star}^{(l)}\Sigma_{\star}^{(l)} {V_{\star}^{(l)}}^\top \\
    &= \sum_{i=1}^{r} u_{\star}^{(l, i)} \sigma_{\star}^{(l, i)} v_{\star}^{(l, i)}
\end{aligned}
\label{equ:svd}
\end{equation}
where $\Sigma_{\star}^{(l)}=\text{diag}(\boldsymbol{\sigma})$ with $\boldsymbol{\sigma}=[\sigma_{\star}^{(l, 1)}, \sigma_{\star}^{(l, 2)}, ..., \sigma_{\star}^{(l, r)}]$. $U_{\star}^{(l)}=[u_{\star}^{(l, 1)}, u_{\star}^{(l, 2)}, ..., u_{\star}^{(l, r)}]\in \mathbb{R}^{m_1\times r}$, and $V_{\star}^{(l)}=[v_{\star}^{(l, 1)}, v_{\star}^{(l, 2)}, ..., v_{\star}^{(l, r)}]^\top\in \mathbb{R}^{r\times m_2}$. The rank $r$ and dimensions $m_1$ and $m_2$ are associated with $W_{\star}^{(l)}$. Each basic component is represented as $\Theta_{\star}^{(l, i)} = (u_{\star}^{(l, i)}, \sigma_{\star}^{(l, i)}, v_{\star}^{(l, i)})$.

\subsection{Knowledge Diversion by Class Labels}
\label{sec:know div}
Given a dataset with $N_{cls}$ classes, our objective is to allocate knowledge of each class to the corresponding basic components while extracting class-agnostic knowledge shared across all classes, thereby achieving knowledge diversion.

We categorize all basic components into \textbf{\textit{learngenes}} and \textbf{\textit{tailors}}, encapsulating class-agnostic and class-specific knowledge, respectively.  
Specifically, the components are partitioned based on the number of classes $N_{cls}$ and matrix rank $r$, satisfying $r = N_{cls} \cdot N_T + N_{G}$, where $N_T$ denotes the number of components per class, with the tailor for the $c$-th class $\mathcal{T}_c$:
\begin{equation}
    \mathcal{T}_c=\{\Theta_{\star}^{(l, i)}|i\in [(c-1)\cdot N_T, c\cdot N_T], \star \in \mathcal{S},\, l \in [1, L] \}
\end{equation}
$N_G$ is the number of basic components forming learngenes:
\begin{equation}
    \mathcal{G}=\{\Theta_{\star}^{(l, i)}|i\in [N_{cls}\cdot N_T, N_{cls}\cdot N_T + N_G], \star\in \mathcal{S}, l\in [1, \small{L}]\}
\end{equation}
In this way, the $r$ basic components of each matrix are partitioned into $N_G$ learngenes and $N_{cls}$ tailors, with the model parameters represented as $\theta = \mathcal{G} + \sum_{c=1}^{N_{cls}} \mathcal{T}_c$.

To encapsulate the class-specific knowledge of the $c$-th class in the $c$-th tailor, we introduce a class gate $G = [0, \ldots, 0, 1, 0, \ldots, 0] \in \mathbb{R}^{N_{cls}}$ for knowledge diversion during the training of DiTs, where only one the element at the $c$-th position is set to 1, corresponding to the class index. 
This mechanism ensures that, for each training class, only the weight parameters of the learngene and relevant tailors are updated (See Algorithm~\ref{alg:algorithm} for more details). The optimization objective is defined as:
\begin{equation}
    {\underset{\mathcal{G}, \mathcal{T}}{{\arg\min}} \, \mathcal{L}_{G\cdot \theta}},\;\;\text{s.t.}\, \theta = \mathcal{G} + \sum_{c=1}^{N_{cls}} \mathcal{T}_c
\label{equ:ob}
\end{equation}
where the loss function $\mathcal{L}$ is defined in Eq.~\eqref{eq:loss}.

\subsection{Decomposable Models for Diverse Scenarios}
\begin{figure}[tb]
  \centering
  \includegraphics[width=\linewidth]{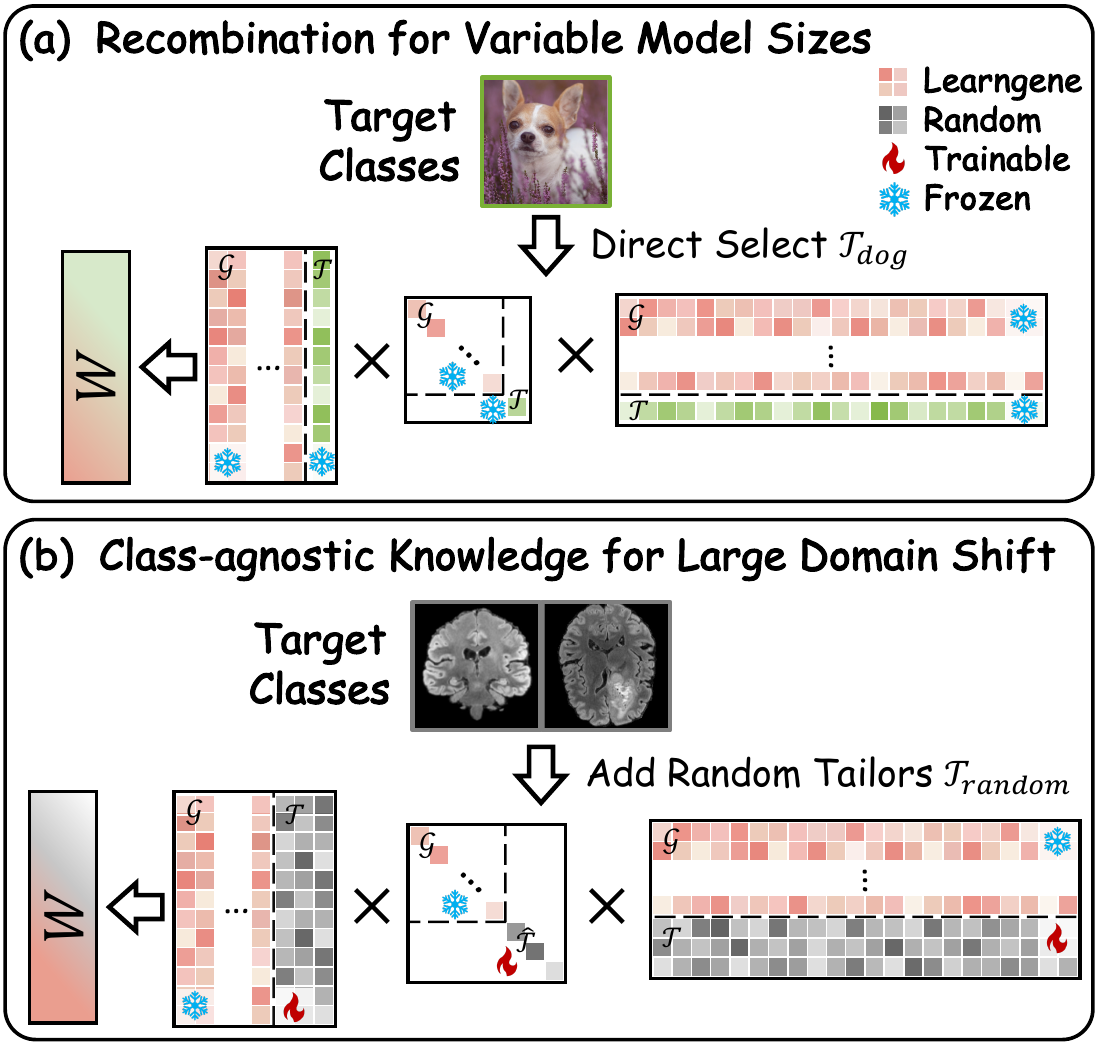}
  \vspace{-0.25in}
  \caption{(a) For downstream tasks with pre-trained classes, it can directly select the tailors corresponding to the target classes while discarding unrelated ones. (b) When encountering tasks with large domain shifts, only the learngene is transferred, combined with randomly initialized tailors for class-specific fine-tuning.}
  \label{fig:inherit}
  \vspace{-0.15in}
\end{figure}
After training via knowledge diversion, we obtain a decomposable model made up of basic components, which can be adaptively reassembled to meet the target memory size and specific task requirements during deployment.

\textbf{Recombination for Variable Model Sizes.} 
In practice, not all knowledge in pre-trained models is applicable to downstream tasks, and transferring excessive knowledge can be both memory-intensive and redundant.
For downstream tasks similar to parts of the training dataset, we can directly select the appropriate pre-trained tailors combined with learngenes. 
For instance, when deploying a DiT pre-trained on \textbf{\textit{ImageNet}} to a resource-constrained device for generating images of \textit{\textbf{``dogs''}}, we can deploy only the tailor corresponding to ``dog'' ($\mathcal{T}_{\text{dog}}$) and the learngene ($\mathcal{G}$). 
Similarly, for unknown classes, we can select closely related tailors for fine-tuning, adjusting the number of tailors based on the available memory.

\textbf{Class-agnostic Knowledge for Large Domain Shift.}
Pre-trained models often encounter negative transfer when facing large domain shifts, a challenge that also affects the transfer of pre-trained tailors.
In such cases, class-agnostic knowledge encapsulated in learngenes fully demonstrates its advantages.
Thus, for tasks with large domain shifts, only learngenes need to be transferred, along with randomly initialized tailors $\mathcal{T}_{\text{random}}$. 
During fine-tuning, we freeze the learngene and only update the tailors, enabling them to learn class-specific knowledge from the downstream task, thereby achieving more efficient fine-tuning.

\section{Experiments}
\subsection{Datasets}
We conduct class-conditioned generation on ImageNet-1K~\cite{deng2009imagenet}, which contains 1,000 classes. 
To minimize inter-class similarity, we merge certain similar classes based on their superclasses in WordNet~\cite{miller1995wordnet}, resulting in a final set of 611 classes. 
Among these, 150 classes are used for pre-training the diffusion models, while the remaining 461 classes serve as novel classes for constructing downstream tasks. Further details can be found in Appendix~\ref{sec:dataset}. 
Additionally, we use datasets, including CelebA-HQ~\cite{huang2018introvae}, Hubble~\cite{weinzierl2023hubble}, MRI, and Pokémon, to simulate large domain shifts compared to the training data.

\subsection{Basic Setting}
For pre-training DiT, we train class-conditional latent DiTs of sizes -B and -L, with a latent patch size of $p=2$ at a $256\times 256$ image resolution on training classes. 
All models are trained using AdamW with a batch size of 256 and a constant learning rate of $1\times10^{-4}$ over 300K steps.  
An exponential moving average (EMA) of DiT weights is used with a decay rate of 0.9999, and results are reported using the EMA model.
During image generation, a classifier-free guidance (cfg) scale of 1.5 is applied. Performance is evaluated using Fr\'{e}chet Inception Distance (FID)~\cite{heusel2017gans}, sFID~\cite{nash2021generating}, Fr\'{e}chet DINO distance(FDD)~\cite{stein2023exposing}, Inception Score~\cite{salimans2016improved} and Precision/Recall~\cite{kynkaanniemi2019improved}. Further details are provided in Appendix~\ref{sec:hyper}.

\renewcommand{\arraystretch}{0.9}
\begin{table}
    \centering
    \setlength{\tabcolsep}{1.0 mm}
    \caption{Performance of constructing variable-sized models on training classes. ``Para.'' denotes the total number of model parameters, which reflects the model size. ``Time'' is the additional training steps required to construct models of the target sizes.}
    \vspace{0.02in}
        \resizebox{0.49\textwidth}{!}{
        \begin{tabular}{@{}lcccccccc@{}}
        \toprule[1.5pt]
        & Para.\small{(M)} & Methods & Time & FID$\downarrow$ & sFID$\downarrow$ & IS$\uparrow$ & Prec.$\uparrow$ & Rec.$\uparrow$\\
        \midrule[1.1pt]    

        \multirow{5}{*}{\rotatebox{90}{\textbf{DiT-L}}}
        & \cellcolor{red!10}{457.0} & \cellcolor{red!10}{Trad. PT} & \cellcolor{red!10}{0} & \cellcolor{red!10}{9.68} & \cellcolor{red!10}{\textbf{6.15}} & \cellcolor{red!10}{72.22} & \cellcolor{red!10}{0.69} & \cellcolor{red!10}{\textbf{0.47}} \\
        & 362.5 & Heur-LG & 100K & 23.86 & 7.24 & 48.34 & 0.54 & 0.47 \\
        & 249.2 & Laptop-diff & 100K & 17.20 & 7.25 & 57.07 & 0.59 & 0.47 \\
        & 249.2 & Auto-LG & 100K & 18.38 & 8.22 & 57.68 & 0.58 & 0.46 \\
        & \cellcolor{blue!12}{\textbf{245.9}} & \cellcolor{blue!12}{KIND} & \cellcolor{blue!12}{\textbf{0}} & \cellcolor{blue!12}{\textbf{9.33}} & \cellcolor{blue!12}{6.80} & \cellcolor{blue!12}{\textbf{79.39}} & \cellcolor{blue!12}{\textbf{0.69}} & \cellcolor{blue!12}{0.46} \\
        \midrule[1.1pt]
        \multirow{5}{*}{\rotatebox{90}{\textbf{DiT-B}}}
        & \cellcolor{red!10}{129.7} & \cellcolor{red!10}{Trad. PT} & \cellcolor{red!10}{0} & \cellcolor{red!10}{25.14} & \cellcolor{red!10}{\textbf{7.57}} & \cellcolor{red!10}{47.15} & \cellcolor{red!10}{0.53} & \cellcolor{red!10}{0.46} \\
        & 108.4 & Heur-LG & 100K & 41.53 & 8.93 & 34.29 & 0.42 & 0.47 \\
        & 76.5 & Laptop-diff & 100K & 48.22 & 11.09 & 31.19 & 0.37 & \textbf{0.47} \\
        & 76.5 & Auto-LG & 100K & 45.69 & 10.77 & 32.77 & 0.39 & 0.47 \\
        & \cellcolor{blue!12}{\textbf{70.2}} & \cellcolor{blue!12}{KIND} & \cellcolor{blue!12}{\textbf{0}} & \cellcolor{blue!12}{\textbf{21.14}} & \cellcolor{blue!12}{8.85} & \cellcolor{blue!12}{\textbf{58.18}} & \cellcolor{blue!12}{\textbf{0.55}} & \cellcolor{blue!12}{0.44} \\

        \bottomrule[1.5pt]
        \end{tabular}
        }
    \label{tab:re_con}
    \vspace{-0.15in}
\end{table}

\begin{table*}
    \centering
    \setlength{\tabcolsep}{1 mm}
    \caption{Performance of various PEFT and learngene methods on novel classes. All methods are fine-tuned for 50K steps on 18 downstream tasks involving novel classes. “Para.” denotes the average number of trainable parameters, while ``FLOPs'' represents the average total floating-point operations required during fine-tuning.}
    \vspace{0.02in}
    \resizebox{\textwidth}{!}{
        \begin{tabular}{@{}llccccccc|ccccccc@{}}
        \toprule[1.5pt]
        \multicolumn{2}{l}{\multirow{2}{*}{Methods}} & \multicolumn{7}{c|}{DiT-B/2} & \multicolumn{7}{c}{DiT-L/2}\\ 
        \cmidrule{3-16}
        & & Para.\small{(M)} & FLOPs\small{(G)} & FID$\downarrow$ & sFID$\downarrow$ & IS$\uparrow$ & Prec.$\uparrow$ & Recall$\uparrow$ & Para.\small{(M)} & FLOPs\small{(G)} & FID$\downarrow$ & sFID$\downarrow$ & IS$\uparrow$ & Prec.$\uparrow$ & Recall$\uparrow$\\
        \midrule[1.1pt]    
        \multirow{7}{*}{\rotatebox{90}{\textbf{PEFT}}}
        & SVDiff 
        & \textit{\textbf{0.1}} & \textit{43.6} & 55.01 & 18.12 & 19.6 & 0.35 & 0.55
        & \textit{\textbf{0.2}} & \textit{155.0} & 49.59 & 16.81 & 20.8 & 0.38 & 0.56\\
        & OFT 
        & \textit{14.2} & \textit{119.7} & 36.19 & 17.79 & 32.0  & 0.48 & 0.50
        & \textit{50.5} & \textit{425.6} & 24.81 & 18.27 & 44.1 & 0.59 & 0.47\\
        & LoRA 
        & \textit{12.8} & \textit{50.1} & 36.70 & 16.28 &  31.6 & 0.44 & 0.57
        & \textit{45.3} & \textit{178.2} & 22.55 & 14.00 & 46.3 & 0.55 & 0.56\\
        & PiSSA 
        & \textit{12.8} & \textit{50.1} & 33.16 & 15.51 & 34.6  & 0.49 & 0.52
        & \textit{45.3} & \textit{178.2} & 19.41 & 14.72 & 53.7 & 0.63 & 0.50\\
        & LoHa 
        & \textit{12.7} & \textit{87.1} & 42.38 & 17.37 & 27.3 & 0.40 & \textbf{0.58}
        & \textit{45.3} & \textit{309.6} & 29.79 & 15.17 & 35.8 & 0.49 & \textbf{0.59}\\
        & DoRA
        & \textit{12.8} & \textit{129.5} & 35.87 & 16.40 & 32.3 & 0.45 & 0.56
        & \textit{45.6} & \textit{503.0} & 21.28 & 14.16 & 48.3 & 0.57 & 0.55\\
        \midrule
        \multirow{3}{*}{\rotatebox{90}{\textbf{LG}}}& Heur-LG 
        & \textit{129.6} & \textit{43.6} & 55.45 & 22.14 & 24.4 & 0.33 & 0.48
        & \textit{456.8} & \textit{155.0} & 41.83 & 19.23 & 30.9 & 0.40 & 0.51\\
        & Auto-LG 
        & \textit{129.6} & \textit{43.6} & 56.38 & 21.39 & 25.5 & 0.30 & 0.49
        & \textit{456.8} & \textit{155.0} & 31.78 & 18.71 & 41.7 & 0.46 & 0.54\\
        & \cellcolor{blue!12}{KIND} 
        & \cellcolor{blue!12}{\textit{12.8}} & \cellcolor{blue!12}{\textit{\textbf{33.7}}} & \cellcolor{blue!12}{\textbf{20.94}} & \cellcolor{blue!12}{\textbf{14.75}} & \cellcolor{blue!12}{\textbf{62.4}} & \cellcolor{blue!12}{\textbf{0.53}} & \cellcolor{blue!12}{0.50}
        & \cellcolor{blue!12}{\textit{45.4}} & \cellcolor{blue!12}{\textit{\textbf{119.6}}} & \cellcolor{blue!12}{\textbf{12.87}} & \cellcolor{blue!12}{\textbf{12.93}} & \cellcolor{blue!12}{\textbf{86.1}} & \cellcolor{blue!12}{\textbf{0.65}} & \cellcolor{blue!12}{0.51}\\
        \midrule
        \multirow{1}{*}{\rotatebox{90}{\textbf{FT}}} & Full FT 
        & \textit{129.6} & \textit{43.6} & 26.49 & 15.08 & 45.1 & 0.51 & 0.55
        & \textit{456.8} & \textit{155.0} & 14.51 & 13.16 & 69.1 & 0.63 & 0.55\\
        \bottomrule[1.5pt]
        \end{tabular}
        }
    \label{tab:main}
    \vspace{-0.15in}
\end{table*}

\begin{figure*}[tb]
  \centering
  \includegraphics[width=\linewidth]{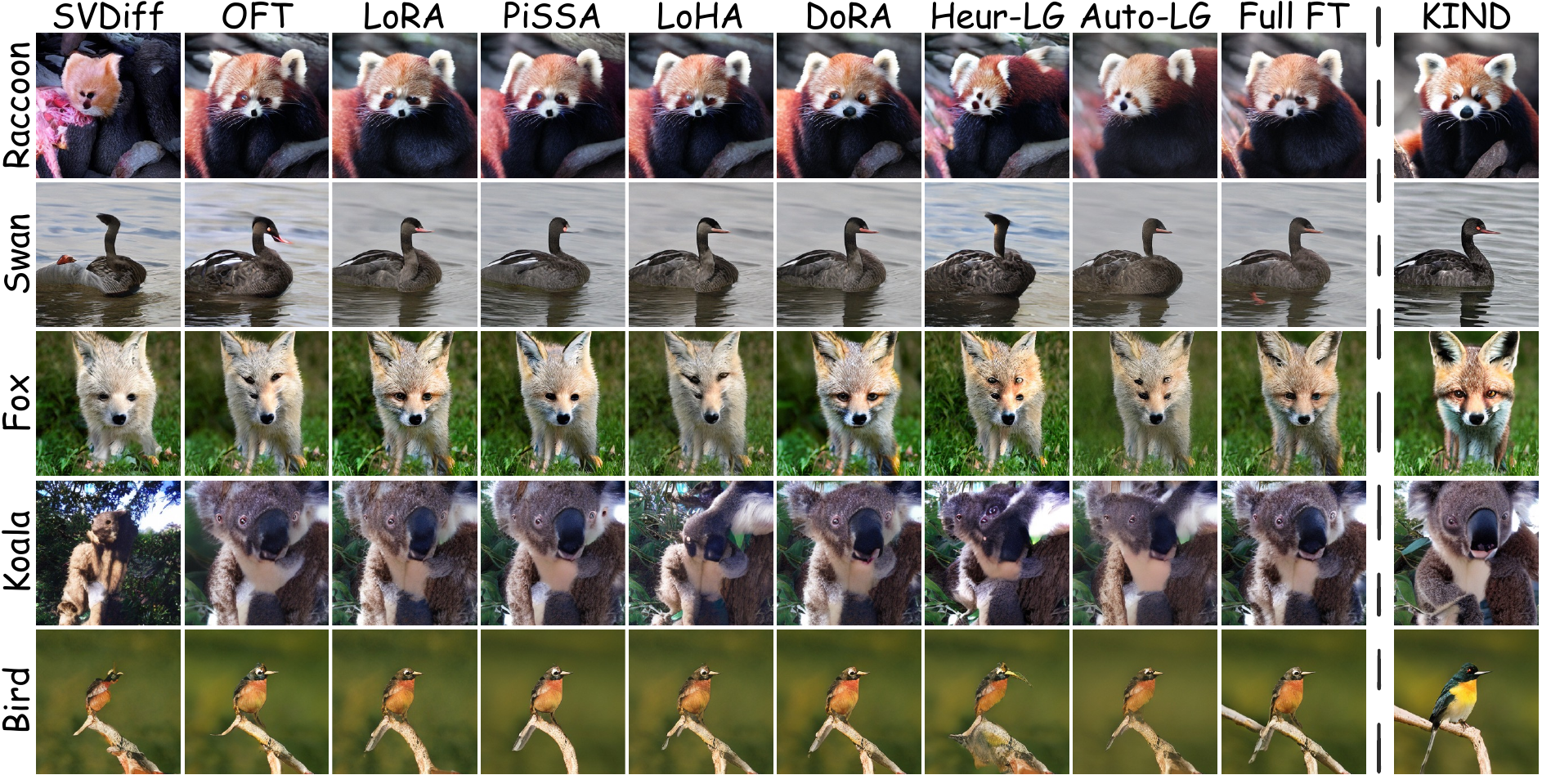}
  \vspace{-0.3in}
  \caption{Selected samples from tasks with novel classes, generated by KIND and other PEFT methods using the DiT-L/2 model, with a resolution of $256 \times 256$. All images are generated using a classifier-free guidance (cfg) scale of 3.0.}
  \label{fig:main}
  \vspace{0in}
\end{figure*}

\section{Results}
\subsection{Construction of Variable-Sized Pre-Trained Models}
The models pre-trained by KIND are inherently decomposable, consisting of \textit{learngenes} that encapsulate class-agnostic knowledge and \textit{tailors} that capture class-specific knowledge. 
This decomposition enables flexible deployment of models across devices, as demonstrated in Table~\ref{tab:re_con}.

Compared to traditional pre-trained models, KIND achieves comparable performance with the same number of training steps, without increasing training complexity.
Additionally, the decomposable nature of KIND allows for direct recombination tailored to specific deployment needs, with \textbf{\textit{no further time-consuming}} steps required. 
In contrast to knowledge distillation and pruning~\cite{zhang2024laptop}, KIND offers significant advantages by avoiding the resource overhead of \textbf{\textit{repeated distillation and pruning}} for each model size, which is required in distillation-based methods.

Unlike traditional learngenes, such as Heur-LG~\cite{wang2022learngene} and Auto-LG~\cite{wang2023learngene}, which directly transfer certain layers from traditional pre-trained models, KIND encapsulates task-agnostic knowledge into learngenes and retains task-specific knowledge in tailors through knowledge diversion. This enables the direct combination of learngenes and tailors without additional training, ensuring both efficiency and adaptability across tasks.

\subsection{Performance on Tasks with Novel Classes}
To evaluate KIND's adaptability, we use learngenes as the backbone with randomly initialized tailors and compare it to PEFT methods based on traditional pre-trained models on tasks with novel classes.
As shown in Table~\ref{tab:main}, KIND achieves state-of-the-art results on DiT-B and DiT-L, reducing FID by 6.54 and sFID by 1.07, while using only 45.4M parameters and saving 35.4G FLOPs on DiT-L.

Despite the efficiency of PEFT methods, a significant performance gap remains compared to Full FT, highlighting the task discrepancy between training and novel classes.
PEFT methods, which freeze pre-trained parameters, struggle to adapt to novel tasks. As shown in Figure~\ref{fig:main}, PEFT-generated images perform poorly in capturing class-specific knowledge due to limited trainable parameters and task mismatch.
Existing learngene methods like Heur-LG and Auto-LG transfer partial knowledge from pre-trained models, but the transferability of each module, trained with traditional objectives, is limited.

In contrast, KIND diverts class-agnostic knowledge into learngenes, creating a flexible backbone for adaptation to downstream tasks with novel classes. The randomly initialized tailors are adjusted via low-rank assumptions, combining with learngenes to meet task-specific needs, thereby improving transfer efficiency and enhancing the generalizability of knowledge transfer. As shown in Figure~\ref{fig:main} and Table~\ref{tab:main}, KIND-generated images outperform PEFT methods in both quality and performance metrics.

\subsection{Performance on Tasks with Large Domain Shifts}
\renewcommand{\arraystretch}{0.95}
\begin{table}
    \centering
    \setlength{\tabcolsep}{1.6 mm}
    \vspace{-0.1in}
    \caption{Performance comparison of KIND and PEFT methods in transferring to downstream tasks with significant domain shifts, evaluated using FDD for image quality assessment.}
    \vspace{0.05in}
    \resizebox{0.49\textwidth}{!}{
        \begin{tabular}{@{}lcc|cc|cc|cc@{}}
        \toprule[1.5pt]
             & \multicolumn{2}{c|}{CelebA-HQ} & \multicolumn{2}{c|}{Hubble} & \multicolumn{2}{c|}{MRI} & \multicolumn{2}{c}{Pokemon}\\
             \cmidrule{2-9}
             & DiT-B & DiT-L & DiT-B & DiT-L & DiT-B & DiT-L & DiT-B & DiT-L \\
             \midrule[1.1pt]
             SVDiff & 0.622 & 0.388 & 0.385 & 0.305 & 0.187 & 0.148 & 0.605 & 0.469 \\
             OFT & 0.343 & 0.226 & 0.255 & 0.168 & 0.056 & 0.046 & 0.469 & 0.321 \\
             LoRA & 0.284 & 0.197 & 0.232 & 0.142 & 0.061 & 0.056 & 0.412 & 0.285\\
             PiSSA & 0.281 & 0.195 & 0.211 & 0.152 & 0.057 & 0.051 & 0.418 & 0.295\\
             LoHa & 0.336 & 0.268 & 0.252 & 0.189 & 0.065 & 0.130 & 0.439 & 0.316\\
             DoRA & 0.282 & 0.203 & 0.589 & 0.330 & 0.043 & 0.048 & 0.396 & 0.333\\
             \midrule
             \cellcolor{blue!12}{KIND} & \cellcolor{blue!12}{\textbf{0.201}} & \cellcolor{blue!12}{\textbf{0.152}} & \cellcolor{blue!12}{\textbf{0.124}} & \cellcolor{blue!12}{\textbf{0.109}} & \cellcolor{blue!12}{\textbf{0.042}} & \cellcolor{blue!12}{\textbf{0.040}} & \cellcolor{blue!12}{\textbf{0.343}} & \cellcolor{blue!12}{\textbf{0.262}}\\
             \bottomrule[1.5pt]
        \end{tabular}
        }
    \label{tab:downstream_ds}
    \vspace{-0.1in}
\end{table}

\begin{figure}[tb]
  \centering
  \includegraphics[width=\linewidth]{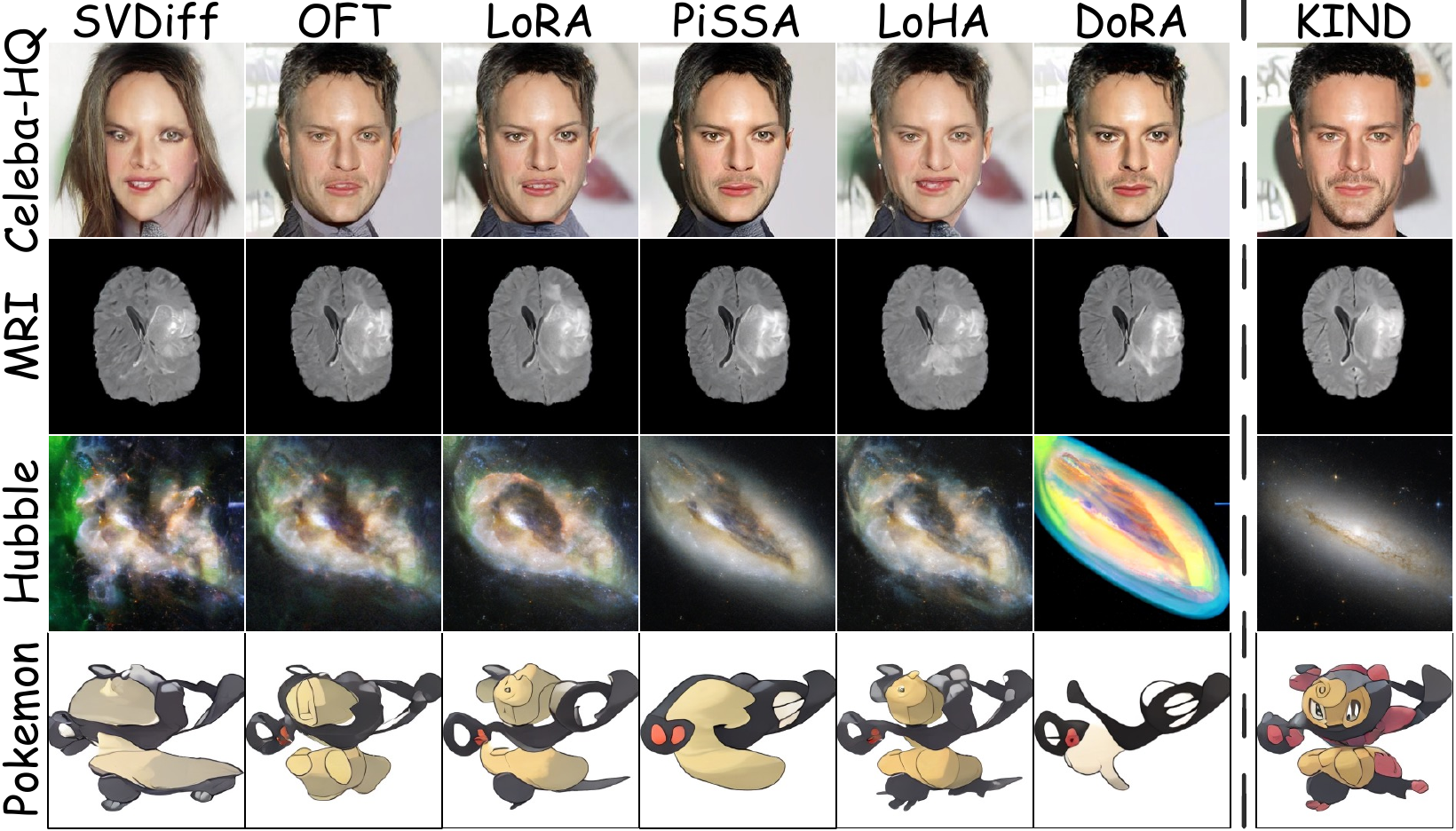}
  \vspace{-0.2in}
  \caption{Selected samples from tasks with large domain shifts, generated by KIND and other PEFT methods using the DiT-L/2, with a resolution of $256 \times 256$. All images are generated using a classifier-free guidance (cfg) scale of 1.5.}  
  \label{fig:main2}
  \vspace{-0.1in}
\end{figure}

KIND demonstrates significant advantages in adapting to tasks with novel classes, with these benefits becoming even more pronounced when dealing with tasks involving large domain shifts. As shown in Table~\ref{tab:downstream_ds} and Figure~\ref{fig:main2}, KIND outperforms PEFT methods on both DiT-B and DiT-L, achieving substantial improvements in image generation quality.

This further demonstrates that the knowledge encapsulated in learngenes is sufficiently class-agnostic, allowing it to be shared effectively across various tasks. In contrast, PEFT methods based on traditional pre-trained models show disadvantages, as the knowledge learned from ImageNet is often difficult to transfer to new domains, especially in specialized fields like Hubble and MRI. This highlights a key limitation of current pre-training approaches, which aim to improve generalization by incorporating as many domain-specific images as possible during training~\cite{ramesh2022hierarchical, esser2024scaling}. While this may enhance performance, it leads to larger model sizes, reduced transfer flexibility, and increased computational overhead.

\subsection{Ablation and Analysis}
\subsubsection{Ablation Experiments}
To assess the effectiveness of learngenes, tailors, and the class gate, we conduct a series of ablation experiments. 
\#1 performs Singular Value Decomposition (SVD) on pre-trained weights and randomly selects $N_G$ singular vectors to form its backbone, followed by fine-tuning with LoRA. \#2 replaces the backbone with learngenes extracted by KIND, based on the structure in \#1. \#3 substitutes tailors for LoRA in fine-tuning the model, without using the class gate.

As shown in Table~\ref{tab:ablation}, the knowledge encapsulated in learngenes, which undergoes knowledge diversion, is more class-agnostic, making it better suited for adaptation to downstream tasks, especially when these tasks differ significantly from the training tasks (e.g., \#1 vs. \#2). 
Additionally, tailors can also function as a PEFT method by integrating class-specific knowledge into pre-trained models or learngenes, thereby enhancing the model's ability to acquire new knowledge for downstream tasks (\#2 vs. \#3). 
Finally, the class gate further enhances this by helping the model distinguish class-specific knowledge, boosting the effectiveness of the tailors (\#3 vs. KIND).
\begin{table}
    \centering
    \setlength{\tabcolsep}{1 mm}
    \vspace{-0.1in}
    \caption{Ablation study on different components of KIND.}
    \vspace{0.05in}
        \resizebox{0.49\textwidth}{!}{
        \begin{tabular}{@{}lccccccccc@{}}
        \toprule[1.5pt]
         & & LG & Tailor & Gate & FID$\downarrow$ & sFID$\downarrow$ & IS$\uparrow$ & Prec.$\uparrow$ & Recall$\uparrow$\\
        \midrule[1.1pt]
        \multirow{4}{*}{\rotatebox{90}{\textbf{DiT-B/2}}}
        & \#1 & & & & 60.28 & 19.96 & 20.4 & 0.30 & 0.49\\
        & \#2 &  \checkmark & & & 49.54 & 18.08 & 23.2 & 0.34 & \textbf{0.56}\\
        & \#3 & \checkmark & \checkmark & & 21.60 & 14.84 & 59.7 & \textbf{0.54} & 0.50 \\
        & \cellcolor{blue!12}{KIND} & \cellcolor{blue!12}{\checkmark} & \cellcolor{blue!12}{\checkmark} & \cellcolor{blue!12}{\checkmark} & \cellcolor{blue!12}{\textbf{20.94}} & \cellcolor{blue!12}{\textbf{14.75}} & \cellcolor{blue!12}{\textbf{62.4}} & \cellcolor{blue!12}{0.53} & \cellcolor{blue!12}{0.50} \\
        \midrule[1.1pt]
        \multirow{4}{*}{\rotatebox{90}{\textbf{DiT-L/2}}}
        & \#1 & & & & 42.04 & 18.07 & 28.0 & 0.41 & 0.54 \\
        & \#2 &  \checkmark & & & 33.53 & 15.55 & 32.2 & 0.46 & \textbf{0.59}\\
        & \#3 & \checkmark & \checkmark & & 13.03 & 12.93 & 85.1 & 0.64 & 0.51 \\
        & \cellcolor{blue!12}{KIND} & \cellcolor{blue!12}{\checkmark} & \cellcolor{blue!12}{\checkmark} & \cellcolor{blue!12}{\checkmark} & \cellcolor{blue!12}{\textbf{12.87}} & \cellcolor{blue!12}{\textbf{12.93}} & \cellcolor{blue!12}{\textbf{86.1}} & \cellcolor{blue!12}{\textbf{0.65}} & \cellcolor{blue!12}{0.51} \\
        \midrule[1.5pt]
        \end{tabular}
        }
    \vspace{-0.15in}
    \label{tab:ablation}
\end{table}

\subsubsection{Strong Learning Ability Brought by Learngenes}
\begin{figure}[tb]
  \centering
  \includegraphics[width=\linewidth]{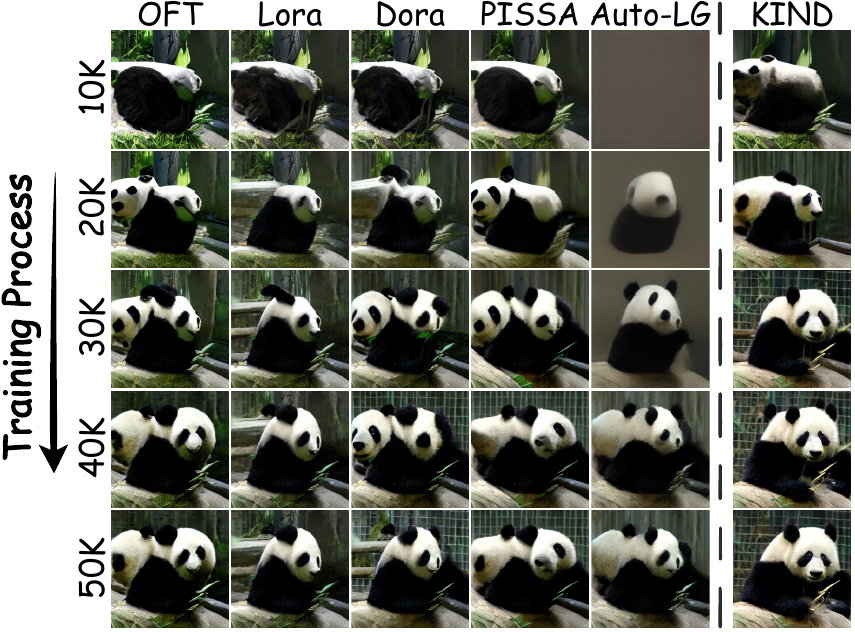}
  \vspace{-0.3in}
  \caption{Visualization of convergence speed of KIND and other methods on downstream tasks. Each image is sampled every 10K steps to illustrate progress more clearly.}
  \vspace{-0.1in}
  \label{fig:fast}
\end{figure}

As noted in \cite{wang2022learngene, xia2024transformer}, learngenes accelerate downstream model adaptation by transferring common knowledge, offering a significant advantage over training from scratch.
Beyond this, KIND further improves convergence speed compared to PEFT methods.
Figure~\ref{fig:fast} illustrates the convergence speed of KIND, with images generated by models every 10K training steps.

The convergence speed is generally influenced by the number of trainable parameters during fine-tuning, with PEFT methods focusing on reducing this number using techniques like orthogonalization and low-rank constraints~\cite{ding2023parameter, han2024parameter}. 
However, these methods often neglect the transferability of knowledge in pre-trained models by directly fixing their parameters.
In contrast, KIND leverages learngenes that encapsulate class-agnostic knowledge as the backbone, offering superior transferability while remaining lightweight. Meanwhile, the tailors capture task-specific knowledge, allowing KIND to achieve faster convergence and improved performance on downstream tasks.

\subsubsection{Analysis on Class-agnostic Knowledge}
\begin{figure}[tb]
  \centering
  \includegraphics[width=\linewidth]{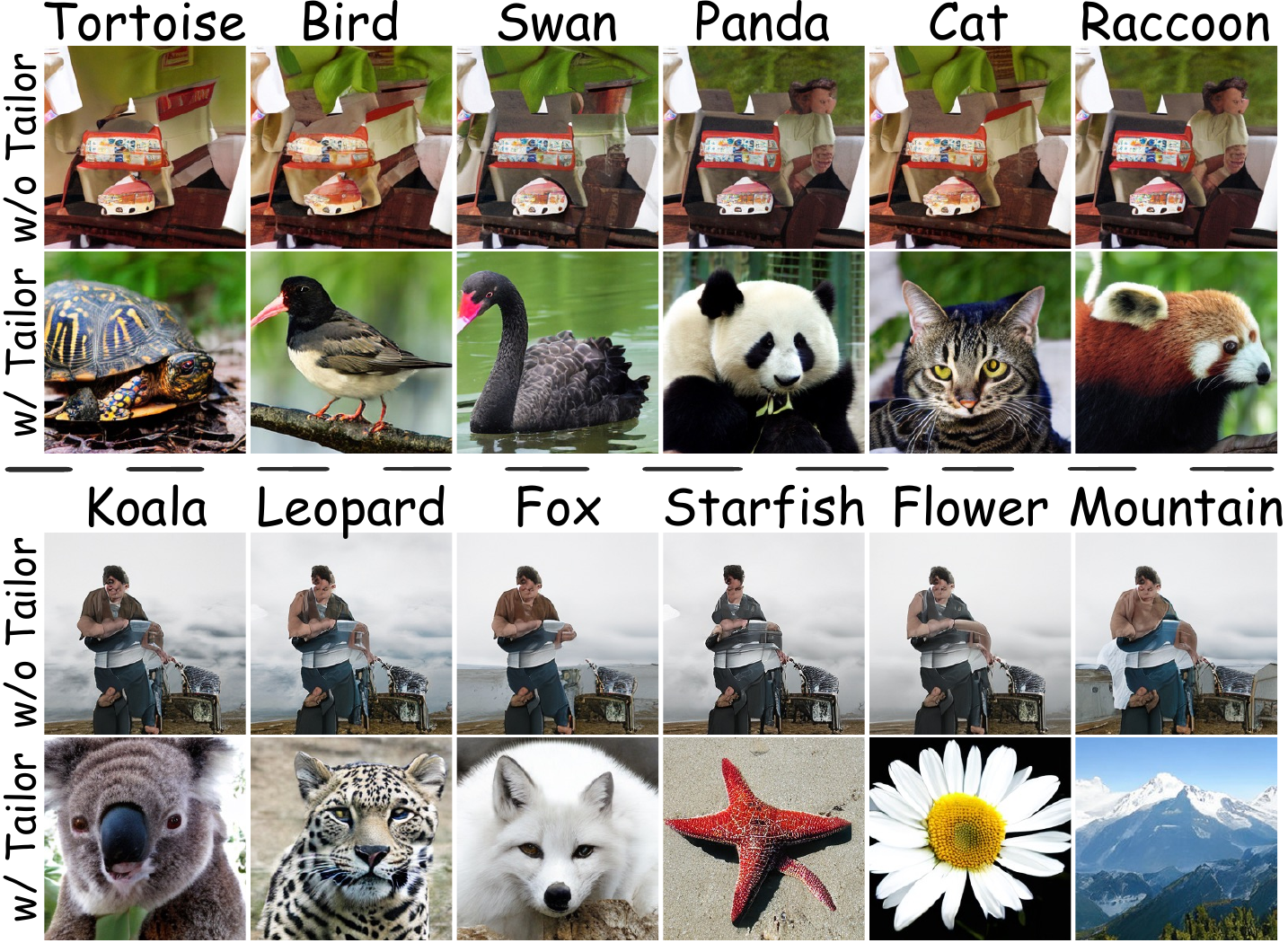}
  \vspace{-0.25in}
  \caption{Visualization of KIND w/ and w/o Tailers (i.e., learngene only) across 12 superclasses for 2 different seeds.}
  \label{fig:common}
  \vspace{-0.1in}
\end{figure}

\begin{table}
    \centering
    \setlength{\tabcolsep}{1.0 mm}
    \caption{Comparison of pre-trained models and learngenes when serving as backbones on training tasks.}
    \vspace{0.05in}
        \begin{tabular}{@{}lccc@{}}
        \toprule[1.5pt]
        & Entropy$\uparrow$ & Variance$\downarrow$ & Kurtosis$\downarrow$ \\
        \midrule[1.1pt]
        Raw Images of ImageNet & 1.458 & 6.414e$^{-4}$ & 884.3  \\
        \midrule
        Pretrained Model & 2.387 & 4.516e$^{-4}$ & 780.1  \\
        \cellcolor{blue!12}{Learngene} & \cellcolor{blue!12}{\textbf{4.046}} & \cellcolor{blue!12}{\textbf{1.495e$\mathbf{^{-4}}$}} & \cellcolor{blue!12}{\textbf{544.9}} \\
        \bottomrule[1.5pt]
        \end{tabular}
    \label{tab:common}
    \vspace{-0.1in}
\end{table}

As discussed earlier, learngenes provide a superior backbone compared to pre-trained models by encapsulating class-agnostic knowledge. 
To further investigate this, we analyze the properties of the class-agnostic knowledge encapsulated in learngenes.
Table~\ref{tab:common} compares learngenes themselves (i.e., w/o tailors) with pre-trained models on training tasks. 
The results reveal that learngenes demonstrate higher entropy, along with lower variance and kurtosis, suggesting that the class-agnostic knowledge they encapsulate is widely applicable across diverse classes. 
Such stability underscores that learngenes, as a backbone, offer better adaptability to unfamiliar classes than traditional pre-trained models.

We also visualize learngenes with and without tailors in Figure~\ref{fig:common}. The visualizations demonstrate that learngenes are not sensitive to category variations, consistently generating similar images across different class conditions. 
While these images may lack detailed semantic information on their own, combining them with class-specific knowledge (i.e., tailors) enables the generation of images corresponding to specific classes. This further underscores the inherent commonality of knowledge within learngenes.

\section{Conclusion}
In this study, we introduce KIND, a pre-training method for constructing decomposable models. KIND employs knowledge diversion during pre-training, separating class-agnostic knowledge into learngenes and class-specific knowledge into tailors. This approach enables the adaptive assembly of variable-sized models by selectively integrating relevant tailors. The class-agnostic knowledge within learngenes mitigates the challenges of tasks with large domain shifts, particularly when combined with randomly initialized tailors for task-specific fine-tuning. We demonstrate the effectiveness of KIND in resource-constrained scenarios and tasks with significant domain shifts, with further analysis and visualizations illustrating the robustness of the class-agnostic knowledge encapsulated in learngenes.

\section*{Acknowledgement}
We sincerely appreciate Freepik for contributing to the figure design. 
This research was supported by the Jiangsu Science Foundation (BK20243012, BG2024036, BK20230832), the National Science Foundation of China (62125602, U24A20324, 92464301, 62306073), China Postdoctoral Science Foundation (2022M720028), and the Xplorer Prize.

\section*{Impact Statement}
The broader impact of our work lies in how KIND redefines the training objectives of pre-trained models, enabling the construction of decomposable models that can be recombined to create models with variable sizes. 
This approach facilitates faster deployment, reduces resource consumption, and enhances adaptability across various tasks and datasets, offering significant value for both research and industrial applications in AI model scaling and transfer learning.

\bibliography{icml2025}
\bibliographystyle{icml2025}

\clearpage
\appendix
\section{Training Details}
\subsection{Details of Knowledge Diversion}
Algorithm \ref{alg:algorithm} presents the pseudo code for diverting class-agnostic knowledge into learngenes and class-specific knowledge into tailors.

\begin{algorithm}[h]
    \caption{Diversion of Class-agnostic Knowledge and Class-specific Knowledge}
    \small
    \label{alg:algorithm}
    \textbf{Input}: DiT $f$, Training dataset $\mathcal{D}=\{(x^{(i)}, y^{(i)})\}_{i=1}^m$ of $N_{cls}$ classes, number of epochs $N_{\text{ep}}$, batch size $B$, learning rate $\alpha$\\
    \textbf{Output}: Learngene $\mathcal{G}$
    \begin{algorithmic}[1]
        \STATE Randomly initialize the weight matrices $\theta$ of $f$, as well as the matrices $U_{\star}^{(l)}$, $\Sigma_{\star}^{(l)}$, and $V_{\star}^{(l)}$
        \FOR{$ep = 1$ to $N_{\text{ep}}$}
            \FOR{each batch $\{(x_i, y_i)\}_{i=1}^B$}
                \STATE Update $\theta$ of $f$ with $U_{\star}^{(l)}$, $\Sigma_{\star}^{(l)}$ and $V_{\star}^{(l)}$ under the rule of Eq.~\eqref{equ:svd}
                \STATE Initialize class gate $G \in \mathbb{R}^{B\times N_{cls}}$ according to labels of images in this batch
                \STATE For each $x_i$, forward propagate $\hat{y}_i = f(x_i, G\cdot \theta)$
                \STATE Calculate $\mathcal{L}_{\text{batch}}=\frac{1}{B} \sum_{i=1}^B \mathcal{L}(\hat{y}_i, y_i)$ according to Eq.~\eqref{eq:loss}
                \STATE Backward propagate the loss $\mathcal{L}_{\text{batch}}$ to compute the gradients with respect to $U_{\star}^{(l)}$, $\Sigma_{\star}^{(l)}$ and $V_{\star}^{(l)}$: $\nabla_{U} \mathcal{L}_{\text{batch}}, \nabla_{\Sigma} \mathcal{L}_{\text{batch}}$ and $\nabla_{V}\mathcal{L}_{\text{batch}}$
                \STATE Update the learngenes $U_{G,\star}^{(l)}$, $\Sigma_{G,\star}^{(l)}$ and $V_{G,\star}^{(l)}$:
                \\ \quad $U_{G,\star}^{(l)} := U_{G,\star}^{(l)} - \alpha \cdot \nabla_{U} \mathcal{L}_{\text{batch}}$, 
                \\ \quad $\Sigma_{G,\star}^{(l)} := \Sigma_{G,\star}^{(l)} - \alpha \cdot \nabla_{\Sigma} \mathcal{L}_{\text{batch}}$
                \\ \quad $V_{G,\star}^{(l)} := V_{G,\star}^{(l)} - \alpha \cdot \nabla_{V} \mathcal{L}_{\text{batch}}$
                \STATE Update the tailors $U_{T_i,\star}^{(l)}$, $\Sigma_{T_i,\star}^{(l)}$ and $V_{T_i,\star}^{(l)}$:
                \\ \quad $U_{T_i,\star}^{(l)} := U_{T_i,\star}^{(l)} - \alpha \cdot G(\nabla_{U} \mathcal{L}_{\text{batch}})$
                \\ \quad $\Sigma_{T_i,\star}^{(l)} := \Sigma_{T_i,\star}^{(l)} - \alpha \cdot G(\nabla_{\Sigma} \mathcal{L}_{\text{batch}})$
                \\ \quad $V_{T_i,\star}^{(l)} := V_{T_i,\star}^{(l)} - \alpha \cdot G(\nabla_{V} \mathcal{L}_{\text{batch}})$
            \ENDFOR
        \ENDFOR
    \end{algorithmic}
\end{algorithm}

\subsection{Hyper-parameters}
\label{sec:hyper}
Table~\ref{tab:hyper_main} presents the basic settings, including learning rate, training steps and the number of learngene components $N_G$ and tailor components $N_T$ for KIND integrating and diverting knowledge.
And Table~\ref{tab:hyper_down} presents the hyper-parameters of PEFT and other learngene methods on 18 downstream tasks. Apart from general hyper-parameters, we also record the hyper-parameters specific to each method. Among them, the parameter $r$ of Lora, PiSSA, Dora and LoHA denotes the rank and the $r$ in OFT denotes the block number respectively.

\begin{table}[tb]
    \centering
    \caption{Hyper-parameters for KIND diverting knowledge on training classes of ImageNet-1K.}
    \vspace{0.05in}
    \setlength{\tabcolsep}{17 mm}
        \begin{tabular}{@{}lr@{}}
        \toprule[1.5pt]
        \textbf{Training Settings} & \textbf{Configuration} \\
        \midrule[1.1pt]
        optimizer & AdamW\\
        learning rate & 1e-4\\
        weight decay & 0\\
        batch size & 256\\
        training steps & 200,000\\
        image size & 256$\times$256\\
        VAE & ema\\
        DiT block & adaLN-Zero\\
        $N_G$ (DiT-B/-L) & 318 / 424\\
        $N_T$ (DiT-B/-L) & 3 / 4\\
        \bottomrule[1.5pt]
        \end{tabular}
    \label{tab:hyper_main}
\end{table}

\begin{table*}
    \centering
    \caption{Hyper-parameters for PEFT and learngene methods when fine-tuning on novel classes of ImageNet-1K.}
    \vspace{0.05in}
    \setlength{\tabcolsep}{1.2 mm}
    \resizebox{\textwidth}{!}{
        \begin{tabular}{@{}lccccccccccccccccccccccc@{}}
        \toprule[1.5pt]
        \textbf{Methods} & \rotatebox{90}{\textbf{Batch Size}} & \rotatebox{90}{\textbf{Training Steps}} & \rotatebox{90}{\makecell[c]{\textbf{Learning Rate}\\(DiT-B / -L)}} & \multicolumn{18}{c}{\textbf{Rank or Block Number $r$}}\\
        & & & & \makecell[c]{Task\\ID} & \#1 & \#2 & \#3 & \#4 & \#5 & \#6 & \#7 & \#8 & \#9 & \#10 & \#11 & \#12 & \#13 & \#14 & \#15 & \#16 & \#17 & \#18\\
        \midrule[1.1pt]
        OFT & 256 & 50K & 1e-4 & & 21 & 11 & 8 & 7 & 6 & 6 & 5 & 5 & 5 & 5 & 5 & 5 & 4 & 4 & 4 & 4 & 4 & 4\\
        Lora & 512 & 50K & 1e-3 & \textbf{-B} & 21 & 39 & 54 & 60 & 69 & 72 & 78 & 78 & 78 & 84 & 84 & 87 & 90 & 90 & 93 & 99 & 102 & 105\\
         & & & & \textbf{-L} & 28 & 52 & 72 & 80 & 92 & 96 & 104 & 104 & 104 & 112 & 112 & 116 & 120 & 120 & 124 & 132 & 136 & 140\\
        PiSSA & 256 & 50K & 1e-3 &  \textbf{-B} & 21 & 39 & 54 & 60 & 69 & 72 & 78 & 78 & 78 & 84 & 84 & 87 & 90 & 90 & 93 & 99 & 102 & 105\\
        & & & & \textbf{-L} & 28 & 52 & 72 & 80 & 92 & 96 & 104 & 104 & 104 & 112 & 112 & 116 & 120 & 120 & 124 & 132 & 136 & 140\\
        Dora & 256 & 50K & 1e-3 & \textbf{-B} & 21 & 39 & 54 & 60 & 69 & 72 & 78 & 78 & 78 & 84 & 84 & 87 & 90 & 90 & 93 & 99 & 102 & 105\\
        & & & & \textbf{-L} & 28 & 52 & 72 & 80 & 92 & 96 & 104 & 104 & 104 & 112 & 112 & 116 & 120 & 120 & 124 & 132 & 136 & 140\\
        LoHA & 256 & 50K & 1e-3 & \textbf{-B} & 10 & 19 & 27 & 30 & 34 & 36 & 39 & 39 & 39 & 42 & 42 & 43 & 45 & 45 & 46 & 49 & 51 & 52\\
        & & & & \textbf{-L} & 14 & 26 & 36 & 40 & 46 & 48 & 52 & 52 & 52 & 56 & 56 & 58 & 60 & 60 & 62 & 66 & 68 & 70\\
        SVDiff & 256 & 50K & 5e-3/3e-3 &  \multicolumn{18}{c}{-----}\\
        Heru-LG & 256 & 50K & 1e-4 & \multicolumn{18}{c}{-----}\\
        Auto-LG & 256 & 50K & 1e-4 & \multicolumn{18}{c}{-----}\\
        KIND & 256 & 50K & 1e-3 & \multicolumn{18}{c}{-----}\\
        Full FT & 256 & 50K & 1e-4 & \multicolumn{18}{c}{-----}\\
        \bottomrule[1.5pt]
        \end{tabular}
        }
    \label{tab:hyper_down}
\end{table*}

\begin{table*}
    \centering
    \caption{Detailed FID of PEFT and learngene methods when fine-tuning on each novel classes.}
    \vspace{0.05in}
    \setlength{\tabcolsep}{0.8 mm}
    \resizebox{0.98\textwidth}{!}{
        \begin{tabular}{@{}l|llcccccccccccccccccccc@{}}
        \toprule[1.5pt]
        \multicolumn{1}{c@{}}{} & \multicolumn{2}{l}{\multirow{2}{*}{\textbf{Methods}}} & \multicolumn{18}{c}{\textbf{Task ID}}\\
        \multicolumn{1}{c@{}}{} & & & \#1 & \#2 & \#3 & \#4 & \#5 & \#6 & \#7 & \#8 & \#9 & \#10 & \#11 & \#12 & \#13 & \#14 & \#15 & \#16 & \#17 & \#18\\
        \midrule[1.1pt]
        \multirow{10}{*}{\rotatebox{90}{\textbf{DiT-B}}} & \multirow{6}{*}{\rotatebox{90}{\textbf{PEFT}}} & SVDiff & 143.4 & 144.9 & 140.6 & 112.3 & 112.5 & 114.2 & 117.4 & 104.3 & 108.6 & 107.5 & 102.3 & 93.6 & 97.5 & 108.9 & 109.6 & 95.2 & 81.6 & 100.6\\
        & & OFT & 92.3 & 90.4 & 93.7 & 71.9 & 76.0 & 86.7 & 82.3 & 72.6 & 74.5 & 76.9 & 65.4 & 63.0 & 67.8 & 77.3 & 78.1 & 64.4 & 63.7 & 75.2\\
        & & Lora & 85.5 & 94.2 & 97.7 & 75.8 & 80.3 & 89.2 & 89.4 & 76.6 & 76.2 & 83.1 & 68.9 & 64.7 & 70.5 & 78.7 & 79.1 & 67.0 & 63.3 & 78.6\\
        & & PiSSA & 83.0 & 89.4 & 93.0 & 69.3 & 73.8 & 82.1 & 81.1 & 69.4 & 71.6 & 76.2 & 64.3 & 60.5 & 64.3 & 74.4 & 70.2 & 60.7 & 59.7 & 71.1\\
        & & LoHa & 94.9 & 100.8 & 108.3 & 84.3 & 88.2 & 95.8 & 97.5 & 85.5 & 86.6 & 90.6 & 78.9 & 73.2 & 79.3 & 88.8 & 88.0 & 76.4 & 69.4 & 86.9\\
        & & Dora & 82.9 & 91.6 & 94.0 & 73.1 & 77.8 & 87.2 & 87.8 & 73.9 & 75.4 & 79.0 & 67.8 & 64.2 & 69.6 & 77.0 & 78.6 & 65.2 & 62.2 & 77.0\\
        \cmidrule{2-21}
        & \multirow{3}{*}{\rotatebox{90}{\textbf{LG}}} & Heru-LG & 98.7 & 111.1 & 122.4 & 97.0 & 102.5 & 114.4 & 122.2 & 95.1 & 99.5 & 108.4 & 87.8 & 90.5 & 91.7 & 103.6 & 101.8 & 94.4 & 88.3 & 100.2\\
        & & Auto-LG & 107.8 & 113.7 & 129.3 & 105.6 & 100.1 & 117.7 & 112.3 & 100.5 & 100.3 & 105.7 & 89.9 & 91.4 & 93.7 & 105.1 & 101.7 & 99.1 & 87.3 & 99.9\\
        & & KIND & \cellcolor{blue!12}{\textbf{55.0}} & \cellcolor{blue!12}{\textbf{73.4}} & \cellcolor{blue!12}{\textbf{70.4}} & \cellcolor{blue!12}{\textbf{52.7}} & \cellcolor{blue!12}{\textbf{58.3}} & \cellcolor{blue!12}{\textbf{65.2}} & \cellcolor{blue!12}{\textbf{59.7}} & \cellcolor{blue!12}{\textbf{47.8}} & \cellcolor{blue!12}{\textbf{51.9}} & \cellcolor{blue!12}{\textbf{56.7}} & \cellcolor{blue!12}{\textbf{42.7}} & \cellcolor{blue!12}{\textbf{43.7}} & \cellcolor{blue!12}{\textbf{44.6}} & \cellcolor{blue!12}{\textbf{56.3}} & \cellcolor{blue!12}{\textbf{62.8}} & \cellcolor{blue!12}{\textbf{43.5}} & \cellcolor{blue!12}{\textbf{39.8}} & \cellcolor{blue!12}{\textbf{52.0}}\\
        \cmidrule{2-21}
        & \multirow{1}{*}{\rotatebox{90}{\textbf{FT}}} & Full FT & 56.3 & 75.5 & 78.1 & 59.9 & 65.1 & 72.6 & 70.1 & 58.6 & 60.1 & 66.1 & 54.2 & 51.4 & 53.8 & 63.7 & 63.3 & 52.8 & 51.5 & 62.7\\
        \midrule[1.5pt]
        \multirow{10}{*}{\rotatebox{90}{\textbf{DiT-L}}} & \multirow{6}{*}{\rotatebox{90}{\textbf{PEFT}}} & SVDiff & 118.2 & 132.0 & 127.2 & 98.3 & 97.2 & 103.0 & 105.6 & 92.3 & 98.3 & 97.2 & 92.9 & 84.1 & 90.5 & 102.3 & 110.4 & 109.0 & 76.3 & 92.3\\
        & & OFT & 59.4 & 71.4 & 72.4 & 52.3 & 57.9 & 65.1 & 64.5 & 55.1 & 60.4 & 58.6 & 48.7 & 50.1 & 52.7 & 62.2 & 60.8 & 50.1 & 51.1 & 61.9\\
        & & Lora & 54.6 & 72.5 & 72.0 & 55.9 & 59.0 & 65.7 & 65.1 & 53.7 & 54.5 & 61.6 & 48.7 & 49.4 & 50.3 & 57.4 & 57.1 & 47.5 & 46.1 & 59.7\\
        & & PiSSA & 52.6 & 68.9 & 67.0 & 50.2 & 54.1 & 60.8 & 58.4 & 49.2 & 48.4 & 55.4 & 43.1 & 44.4 & 44.3 & 53.1 & 48.6 & 41.1 & 41.9 & 50.6\\
        & & LoHa & 65.3 & 78.7 & 83.3 & 63.9 & 69.6 & 77.6 & 78.3 & 66.1 & 66.7 & 73.8 & 62.2 & 59.0 & 62.5 & 68.6 & 68.1 & 59.3 & 56.1 & 72.5\\
        & & Dora & 52.2 & 71.3 & 68.0 & 52.9 & 56.7 & 64.3 & 62.4 & 52.2 & 51.1 & 58.7 & 46.9 & 47.0 & 47.9 & 56.0 & 55.7 & 44.9 & 45.1 & 56.2\\
        \cmidrule{2-21}
        & \multirow{3}{*}{\rotatebox{90}{\textbf{LG}}} & Heru-LG & 73.3 & 92.6 & 97.1 & 79.9 & 86.2 & 94.4 & 94.5 & 77.8 & 82.2 & 88.8 & 72.4 & 71.9 & 77.2 & 85.8 & 85.1 & 74.8 & 71.9 & 84.0\\
        & & Auto-LG & 66.5 & 81.1 & 82.6 & 69.3 & 70.0 & 80.4 & 76.3 & 66.6 & 67.4 & 72.8 & 58.8 & 59.3 & 61.0 & 70.9 & 69.3 & 64.9 & 58.1 & 70.2\\
        & & KIND & \cellcolor{blue!12}{39.0} & \cellcolor{blue!12}{66.2} & \cellcolor{blue!12}{\textbf{61.8}} & \cellcolor{blue!12}{\textbf{44.2}} & \cellcolor{blue!12}{46.0} & \cellcolor{blue!12}{\textbf{54.7}} & \cellcolor{blue!12}{\textbf{47.5}} & \cellcolor{blue!12}{\textbf{39.1}} & \cellcolor{blue!12}{\textbf{40.0}} & \cellcolor{blue!12}{\textbf{46.3}} & \cellcolor{blue!12}{\textbf{33.2}} & \cellcolor{blue!12}{\textbf{36.3}} & \cellcolor{blue!12}{\textbf{34.7}} & \cellcolor{blue!12}{\textbf{45.9}} & \cellcolor{blue!12}{43.9} & \cellcolor{blue!12}{\textbf{31.5}} & \cellcolor{blue!12}{\textbf{30.9}} & \cellcolor{blue!12}{\textbf{40.8}}\\
        \cmidrule{2-21}
        & \multirow{1}{*}{\rotatebox{90}{\textbf{FT}}} & Full FT & \textbf{38.0} & \textbf{64.1} & 61.9 & 44.6 & \textbf{45.5} & 56.1 & 50.8 & 41.4 & 41.2 & 48.5 & 36.1 & 38.6 & 38.2 & 47.4 & \textbf{43.4} & 35.3 & 34.9 & 44.1\\
        \bottomrule[1.5pt]
        \end{tabular}
        }
    \label{tab:down_more}
\end{table*}

\subsection{Details of Downstream Tasks}
\label{sec:dataset}
Table~\ref{tab:datasets} presents the details of 18 downstream tasks, which are sorted by the class numbers in each task. Each task is composed of $c\in [7, 35]$ novel classes, where the classes merged into superclasses in ImageNet1K and their corresponding superclasses are listed in Table \ref{tab:super1} and Table \ref{tab:super2}, while the rest remain the same as the classes in ImageNet-1K.

\renewcommand{\arraystretch}{0.9}
\begin{table*}
    \centering
    \caption{Details of superclasses in each downstream task}
    \vspace{0.05in}
    \setlength{\tabcolsep}{1 mm}
    \resizebox{0.98\textwidth}{!}{
    \begin{tabular}{@{}c|cccccccccc@{}}
        \toprule[1.5pt]
        \textbf{Task} & \multicolumn{9}{c}{\textbf{Superclasses of ImageNet}}\\
        \midrule[1.1pt]
        \textbf{\#1} & n02510455 & n02509815 & n01662784 & n02118333 & n02083346 & n02437616 & n02457408 \\ 
        \midrule
        \textbf{\#2} & n03187595 & n03788365 & n03933933 & n04273569 & n03843555 & n03400231 & n03325584 & n09472597 & n03874293 & n04591713 \\
        & n03854065 & n03868863 & n07711569 \\
        \midrule
        \textbf{\#3} & n07753592 & n03763968 & n03109150 & n09399592 & n03903868 & n03720891 & n02939185 & n03908714 & n04014297 & n02804414 \\
        & n06785654 & n04131690 & n02794156 & n02971356 & n02056570 & n02965783 & n04243546 & n06359193 \\
        \midrule
        \textbf{\#4} & n02877765 & n04238763 & n04009552 & n03666591 & n07614500 & n09332890 & n01629276 & n04483307 & n03291819 & n02120997 \\
        & n03717622 & n04041544 & n03873416 & n04467665 & n03394916 & n03272010 & n04118538 & n04367480 & n04447861 & n03775071 \\
        \midrule
        \textbf{\#5} & n04086273 & n04141076 & n03657121 & n03379051 & n02401031 & n01503061 & n03840681 & n04380533 & n03871628 & n11879895 \\
        & n04090263 & n04557648 & n03016953 & n02808304 & n02879718 & n03724870 & n04423845 & n02917067 & n03691459 & n02672831 \\
        & n04146614 & n04525305 & n04264628 \\
        \midrule
        \textbf{\#6} & n03496892 & n06874185 & n04392985 & n03485794 & n03982430 & n04540053 & n03602883 & n02871525 & n02978881 & n03961711\\
        & n04005630 & n03065424 & n04200800 & n02823750 & n03344393 & n04325704 & n03220513 & n03498962 & n04356056 & n03347037\\
        & n09421951 & n07760859 & n04133789 & n07565083 \\
        \midrule
        \textbf{\#7} &n04332243 & n02883205 & n03405725 & n03017168 & n04553703 & n03777568 & n02951358 & n07720875 & n03637318 & n02090827 \\
        & n04265275 & n03028079 & n07920052 & n03954731 & n04141327 & n03255030 & n03447447 & n00002684 & n03530642 & n03425413 \\
        & n04524313 & n03110669 & n03764736 & n12267677 & n02676566 & n03417042 \\
        \midrule
        \textbf{\#8} & n03676483 & n02865351 & n03792972 & n02974003 & n02906734 & n07860988 & n03249569 & n00021265 & n02727426 & n03782006 \\
        & n02317335 & n02815834 & n03388043 & n03529860 & n02817516 & n03761084 & n09246464 & n03899768 & n03970156 & n04485082 \\
        & n01769347 & n07880968 & n03197337 & n03876231 & n02699494 & n03472232 \\
        \midrule
        \textbf{\#9} & n02121808 & n07734744 & n03424325 & n03494278 & n03935335 & n03690938 & n03240683 & n03467068 & n02980441 & n03450230 \\
        & n02512053 & n04517823 & n02730930 & n03133878 & n03259280 & n04376876 & n03803284 & n03920288 & n02966193 & n02814860 \\
        & n02669723 & n03000134 & n02793495 & n02766320 & n03649909 & n04125021 \\
        \midrule
        \textbf{\#10} & n03985232 & n03590841 & n03388549 & n04065272 & n03633091 & n02916936 & n03201208 & n04208210 & n02988304 & n09229709 \\
        & n02769748 & n02791270 & n03814639 & n03481172 & n03692522 & n04501370 & n03584829 & n02843684 & n04252225 & n03196217 \\
        & n02704792 & n03384352 & n03785016 & n03459775 & n03599486 & n01806143 & n03294048 & n03995372 \\ 
        \midrule
        \textbf{\#11} & n04341686 & n03603722 & n04081281 & n03623198 & n03497657 & n02690373 & n09193705 & n04486054 & n01986214 & n01639765 \\
        & n03180011 & n03532672 & n03540267 & n02356798 & n03662601 & n04277352 & n04204238 & n04204347 & n04530566 & n04033901 \\
        & n03793489 & n02268148 & n04209239 & n04266014 & n01861778 & n03062245 & n03179701 & n11939491 \\ 
        \midrule
        \textbf{\#12} & n04111531 & n04597913 & n07932039 & n04118776 & n02859443 & n04523525 & n02077923 & n03938244 & n07707451 & n04371430 \\
        & n02797295 & n04228054 & n03207743 & n01882714 & n07716906 & n03216828 & n04589890 & n03063689 & n03630383 & n04252077 \\
        & n02153203 & n03207941 & n03908618 & n03796401 & n07697313 & n02898711 & n04548362 & n03290653 & n02930766 \\
        \midrule
        \textbf{\#13} & n03000247 & n04040759 & n04590129 & n03492542 & n03733805 & n04044716 & n01877812 & n04418357 & n09428293 & n03045698 \\
        & n03998194 & n03443371 & n03983396 & n03902125 & n03598930 & n01844917 & n04509417 & n02441326 & n02786058 & n03134739 \\
        & n03838899 & n04192698 & n02837789 & n02074367 & n02701002 & n07717070 & n03977966 & n12992868 & n03445777 & n04162706 \\ 
        \midrule
        \textbf{\#14} & n03538406 & n03314780 & n03916031 & n04310018 & n04074963 & n04462240 & n03250847 & n01704323 & n07753113 & n04532106 \\
        & n09288635 & n04033995 & n03929855 & n03733281 & n04562935 & n03124043
        & n03682487 & n04487081 & n03743016 & n03670208 \\
        & n03980874 & n04596742 & n03457902 & n04536866 & n03085013 & n03527444 & n04099969 & n04141975 & n04326547 & n02825657 \\ 
        \midrule
        \textbf{\#15} & n04417672 & n02966687 & n03868242 & n02692877 & n04435653 & n04039381 & n02084071 & n02776631 & n02950826 & n04350905 \\
        & n04552348 & n07831146 & n04149813 & n03787032 & n03791053 & n04357314 & n04476259 & n02129604 & n03791235 & n03992509 \\
        & n01604330 & n03891332 & n04613696 & n04592741 & n02687172 & n02782093 & n04525038 & n02835271 & n01674464 & n07742313 \\
        & n02454379 \\
        \midrule
        \textbf{\#16} & n02910353 & n02323902 & n03327234 & n01726692 & n03095699 & n04443257 & n04201297 & n02667093 & n04584207 & n04328186 \\
        & n02909870 & n04311174 & n04067472 & n04270147 & n04344873 & n03777754 & n03658185 & n03706229 & n07836838 & n03770679 \\
        & n03208938 & n01976146 & n02062744 & n03697007 & n03476684 & n02469914 & n04458633 & n02274259 & n10565667 & n01872401 \\
        & n03584254 & n04019541 & n03461385 \\
        \midrule
        \textbf{\#17} & n03063599 & n04576211 & n03841143 & n03617480 & n02992211 & n04251144 & n04239074 & n02131653 & n04254120 & n02979186 \\
        & n01514668 & n03476991 & n04229816 & n03776460 & n04429376 & n01696633
        & n01905661 & n03594945 & n04370456 & n02159955 \\
        & n04230808 & n03141823 & n00001930 & n03485407 & n04372370 & n04285008 & n03032252 & n04286575 & n02894605 & n03709823 \\
        & n02329401 & n03160309 & n03721384 & n03857828 \\
        \midrule
        \textbf{\#18} & n02870880 & n03127747 & n02880940 & n04346328 & n04482393 & n03800933 & n04152593 & n03051540 & n03042490 & n04317175 \\
        & n03661043 & n04548280 & n04235860 & n02807133 & n02790996 & n03877472 & n07892512 & n07871810 & n03866082 & n07875152 \\
        & n10148035 & n04531098 & n03814906 & n02927161 & n04296562 & n03729826 & n04023962 & n01768244 & n00003553 & n04127249 \\
        & n04505470 & n03825788 & n03794056 & n03929660 & n03742115\\
        \bottomrule[1.5pt] 
    \end{tabular}
    }
    \label{tab:datasets}
\end{table*}

\begin{table*}
    \centering
    \caption{Details of superclasses in ImageNet-1K}
    \vspace{0.05in}
    \setlength{\tabcolsep}{2 mm}
    \resizebox{0.98\textwidth}{!}{
    \begin{tabular}{@{}c|cccccccc@{}}
        \toprule[1.5pt]
        \textbf{Superclass} & \multicolumn{8}{c}{\textbf{Classes of ImageNet}}\\
        \midrule[1.1pt]
        \textbf{n02084071} & n02085620 & n02085782 & n02085936 & n02086079
                  & n02086240 & n02086646 & n02086910 & n02087046 \\
                  & n02087394 & n02088094 & n02088238 & n02088364 
                  & n02088466 & n02088632 & n02089078 & n02089867 \\
                  & n02089973 & n02090379 & n02090622 & n02090721 
                  & n02091244 & n02091467 & n02091635 & n02091831 \\
                  & n02092002 & n02092339 & n02093256 & n02093428 
                  & n02093647 & n02093754 & n02093859 & n02093991 \\
                  & n02094114 & n02094258 & n02094433 & n02095314 
                  & n02095570 & n02095889 & n02096051 & n02096177 \\
                  & n02096294 & n02096437 & n02096585 & n02097047 
                  & n02097130 & n02097209 & n02097298 & n02097474 \\
                  & n02097658 & n02098105 & n02098286 & n02098413 
                  & n02099267 & n02099429 & n02099601 & n02099712 \\
                  & n02099849 & n02100236 & n02100583 & n02100735 
                  & n02100877 & n02101006 & n02101388 & n02101556 \\
                  & n02102040 & n02102177 & n02102318 & n02102480 
                  & n02102973 & n02104029 & n02104365 & n02105056 \\
                  & n02105162 & n02105251 & n02105412 & n02105505 
                  & n02105641 & n02105855 & n02106030 & n02106166 \\
                  & n02106382 & n02106550 & n02106662 & n02107142 
                  & n02107312 & n02107574 & n02107683 & n02107908 \\
                  & n02108000 & n02108089 & n02108422 & n02108551 
                  & n02108915 & n02109047 & n02109525 & n02109961 \\
                  & n02110063 & n02110185 & n02110341 & n02110627 
                  & n02110806 & n02110958 & n02111129 & n02111277 \\
                  & n02111500 & n02111889 & n02112018 & n02112137 
                  & n02112350 & n02112706 & n02113023 & n02113186 \\
                  & n02113624 & n02113712 & n02113799 & n02113978 \\
        \midrule
        \textbf{n01503061} & n01530575 & n01531178 & n01532829 & n01534433  
                  & n01537544 & n01558993 & n01560419 & n01580077 \\ 
                  & n01582220 & n01592084 & n01601694 & n01608432  
                  & n01817953 & n01818515 & n01819313 & n01820546 \\ 
                  & n01824575 & n01828970 & n01829413 & n01833805  
                  & n01843065 & n01843383 & n02002556 & n02002724 \\ 
                  & n02006656 & n02007558 & n02009229 & n02009912 
                  & n02011460 & n02012849 & n02013706 & n02017213 \\
                  & n02018207 & n02018795 & n02025239 & n02027492 
                  & n02028035 & n02033041 & n02037110 & n02051845 \\
                  & n02058221 \\
        \midrule
        \textbf{n02159955} & n02165105 & n02165456 & n02167151 & n02168699
                  & n02169497 & n02172182 & n02174001 & n02177972 \\
                  & n02190166 & n02206856 & n02219486 & n02226429 
                  & n02229544 & n02231487 & n02233338 & n02236044 \\
                  & n02256656 & n02259212 & n02264363 \\
        \midrule
        \textbf{n02469914} & n02481823 & n02483362 & n02483708 & n02484975 
                  & n02486261 & n02486410 & n02487347 & n02488291 \\
                  & n02488702 & n02489166 & n02490219 & n02492035 
                  & n02492660 & n02493509 & n02493793 & n02494079 \\
                  & n02497673 & n02500267 \\
        \midrule
        \textbf{n01726692} & n01728572 & n01728920 & n01729322 & n01729977 
                  & n01734418 & n01735189 & n01737021 & n01739381 \\
                  & n01740131 & n01742172 & n01744401 & n01748264 
                  & n01749939 & n01751748 & n01753488 & n01755581 \\
                  & n01756291 \\
        \midrule
        \textbf{n02512053} & n01440764 & n01443537 & n01484850 & n01491361 
                  & n01494475 & n01496331 & n01498041 & n02514041 \\
                  & n02526121 & n02536864 & n02606052 & n02607072 
                  & n02640242 & n02641379 & n02643566 & n02655020 \\
        \midrule
        \textbf{n01674464} & n01675722 & n01677366 & n01682714 & n01685808 
                  & n01687978 & n01688243 & n01689811 & n01692333 \\
                  & n01693334 & n01694178 & n01695060 \\
        \midrule
        \textbf{n02401031} & n02403003 & n02408429 & n02410509 & n02412080 
                  & n02415577 & n02417914 & n02422106 & n02422699 \\
                  & n02423022 \\
        \midrule
        \textbf{n01769347} & n01770081 & n01773157 & n01773549 & n01773797 
                  & n01774384 & n01774750 & n01775062 & n01776313 \\
        \midrule
        \textbf{n02083346} & n02114367 & n02114548 & n02114712 & n02114855 
                  & n02115641 & n02115913 & n02116738 & n02117135 \\
        \midrule
        \textbf{n02441326} & n02441942 & n02442845 & n02443114 & n02443484 
                  & n02444819 & n02445715 & n02447366 \\
        \midrule
        \textbf{n12992868} & n12985857 & n12998815 & n13037406 & n13040303 
                  & n13044778 & n13052670 & n13054560 \\
        \midrule
        \textbf{n02153203} & n01795545 & n01796340 & n01797886 & n01798484 
                  & n01806567 & n01807496 \\
        \midrule
        \textbf{n02120997} & n02125311 & n02127052 & n02128385 & n02128757 
                  & n02128925 & n02130308 \\
        \midrule
        \textbf{n02274259} & n02276258 & n02277742 & n02279972 & n02280649 
                  & n02281406 & n02281787 \\
        \midrule
        \textbf{n04531098} & n02795169 & n02808440 & n03950228 & n04049303 
                  & n04398044 & n04493381 \\
        \midrule
        \textbf{n01629276} & n01629819 & n01630670 & n01631663 & n01632458 
                  & n01632777 \\
        \midrule
        \textbf{n01662784} & n01664065 & n01665541 & n01667114 & n01667778
                  & n01669191 \\
        \midrule
        \textbf{n01905661} & n01924916 & n01950731 & n01955084 & n01990800 
                  & n02321529 \\
        \midrule
        \textbf{n02121808} & n02123045 & n02123159 & n02123394 & n02123597 
                  & n02124075 \\
        \midrule
        \textbf{n02329401} & n02342885 & n02346627 & n02361337 & n02363005 
                  & n02364673 \\
        \midrule
        \textbf{n04341686} & n03781244 & n03788195 & n03837869 & n03877845 
                  & n03956157 \\
        \bottomrule[1.5pt] 
    \end{tabular}
    }
    \label{tab:super1}
\end{table*}

\begin{table*}
    \centering
    \caption{Details of superclasses in ImageNet-1K (continued)}
    \vspace{0.05in}
    \setlength{\tabcolsep}{3.1 mm}
    \resizebox{0.98\textwidth}{!}{
    \begin{tabular}{@{}c|ccccc|cc@{}}
        \toprule[1.5pt]
        \textbf{Superclass} & \multicolumn{4}{c}{\textbf{Classes of ImageNet}} & \textbf{Superclass} & \multicolumn{2}{c}{\textbf{Classes of ImageNet}}\\
        \cmidrule[1.1pt](r){1-5}
        \cmidrule[1.1pt](l){6-8}
        \textbf{n01976957} & n01978287 & n01978455 & n01980166 & n01981276 
        & \textbf{n02134971} & n02137549 & n02138441 \\
        \cmidrule(r){1-5}
        \cmidrule(l){6-8}  
        \textbf{n02118333} & n02119022 & n02119789 & n02120079 & n02120505
        & \textbf{n02268148} & n02268443 & n02268853 \\
        \cmidrule(r){1-5}
        \cmidrule(l){6-8}  
        \textbf{n02131653} & n02132136 & n02133161 & n02134084 & n02134418 
        & \textbf{n03906997} & n02783161 & n03388183 \\
        \cmidrule(r){1-5}
        \cmidrule(l){6-8}  
        \textbf{n04530566} & n02981792 & n03947888 & n04147183 & n04612504 
        & \textbf{n01604330} & n01614925 & n01616318 \\
        \cmidrule(r){1-5}
        \cmidrule(l){6-8}  
        \textbf{n00021265} & n07579787 & n07583066 & n07584110 & n07590611 
        & \textbf{n01696633} & n01697457 & n01698640 \\
        \cmidrule(r){1-5}
        \cmidrule(l){6-8}  
        \textbf{n01639765} & n01641577 & n01644373 & n01644900 & 
        & \textbf{n01940736} & n01943899 & n01968897 \\
        \cmidrule(r){1-5}
        \cmidrule(l){6-8}  
        \textbf{n01844917} & n01855032 & n01855672 & n01860187 &
        & \textbf{n01942177} & n01944390 & n01945685 \\
        \cmidrule(r){1-5}
        \cmidrule(l){6-8}  
        \textbf{n01861778} & n01871265 & n02504013 & n02504458 &
        & \textbf{n02062744} & n02066245 & n02071294 \\
        \cmidrule(r){1-5}
        \cmidrule(l){6-8}  
        \textbf{n00002684} & n01914609 & n01917289 & n09256479 &
        & \textbf{n02090827} & n02091032 & n02091134 \\
        \cmidrule(r){1-5}
        \cmidrule(l){6-8}  
        \textbf{n01976146} & n01983481 & n01984695 & n01985128 &
        & \textbf{n02134971} & n02137549 & n02138441 \\
        \cmidrule(r){1-5}
        \cmidrule(l){6-8}  
        \textbf{n02323902} & n02325366 & n02326432 & n02328150 &
        & \textbf{n02268148} & n02268443 & n02268853 \\
        \cmidrule(r){1-5}
        \cmidrule(l){6-8}  
        \textbf{n02395003} & n02395406 & n02396427 & n02397096 &
        & \textbf{n03906997} & n02783161 & n03388183\\
        \cmidrule(r){1-5}
        \cmidrule(l){6-8}  
        \textbf{n03472232} & n02777292 & n03535780 & n03888605 &
        & \textbf{n03001627} & n02791124 & n03376595\\
        \cmidrule(r){1-5}
        \cmidrule(l){6-8}  
        \textbf{n03800933} & n02787622 & n02804610 & n03884397 &
        & \textbf{n00001930} & n02799071 & n09835506\\
        \cmidrule(r){1-5}
        \cmidrule(l){6-8}  
        \textbf{n03791235} & n02814533 & n03100240 & n03930630 &
        & \textbf{n04235291} & n02860847 & n03218198\\
        \cmidrule(r){1-5}
        \cmidrule(l){6-8}  
        \textbf{n03497657} & n02869837 & n03124170 & n04259630 &
        & \textbf{n04014297} & n02895154 & n03146219\\
        \cmidrule(r){1-5}
        \cmidrule(l){6-8}  
        \textbf{n03405725} & n03018349 & n03337140 & n04550184 &
        & \textbf{n02883344} & n03014705 & n03127925\\
        \cmidrule(r){1-5}
        \cmidrule(l){6-8}  
        \textbf{n04576211} & n03272562 & n03393912 & n03895866 &
        & \textbf{n03540267} & n03026506 & n04254777\\
        \cmidrule(r){1-5}
        \cmidrule(l){6-8}  
        \textbf{n04230808} & n03534580 & n03770439 & n04136333 &
        & \textbf{n03380867} & n03047690 & n03680355 \\
        \cmidrule(r){1-5}
        \cmidrule(l){6-8}  
        \textbf{n02898711} & n04311004 & n04366367 & n04532670 &
        & \textbf{n03682487} & n03075370 & n03874599 \\
        \cmidrule(r){1-5}
        \cmidrule(l){6-8}  
        \textbf{n07707451} & n07714571 & n07716358 & n07718747 &
        & \textbf{n02766320} & n03125729 & n03131574 \\
        \cmidrule(r){1-5}
        \cmidrule(l){6-8}  
        \textbf{n01604330} & n01614925 & n01616318 & & 
        & \textbf{n03928116} & n03452741 & n04515003 \\
        \cmidrule(r){1-5}
        \cmidrule(l){6-8}  
        \textbf{n01696633} & n01697457 & n01698640 & & 
        & \textbf{n04464852} & n03478589 & n04389033 \\
        \cmidrule(r){1-5}
        \cmidrule(l){6-8}  
        \textbf{n01940736} & n01943899 & n01968897 & & 
        & \textbf{n03985232} & n03642806 & n03832673\\
        \cmidrule(r){1-5}
        \cmidrule(l){6-8}  
        \textbf{n01942177} & n01944390 & n01945685 & & 
        & \textbf{n04524313} & n03673027 & n04347754\\
        \cmidrule(r){1-5}
        \cmidrule(l){6-8}  
        \textbf{n02062744} & n02066245 & n02071294 & & 
        & \textbf{n03051540} & n03710637 & n03710721\\
        \cmidrule(r){1-5}
        \cmidrule(l){6-8}  
        \textbf{n02090827} & n02091032 & n02091134 & &
        & \textbf{n04565375} & n03773504 & n04008634\\
        \cmidrule(r){1-5}
        \cmidrule(l){6-8}  
        \textbf{n02880940} & n03775546 & n04263257 & & 
        & \textbf{n03294048} & n03924679 & n04004767\\
        \cmidrule(r){1-5}
        \cmidrule(l){6-8}  
        \textbf{n03327234} & n03930313 & n04604644 & & 
        & \textbf{n02942699} & n03976467 & n04069434 \\
        \cmidrule(r){1-5}
        \cmidrule(l){6-8}  
        \textbf{n03603722} & n04560804 & n04579145 & & 
        & \textbf{n07679356} & n07684084 & n07695742 \\
        \cmidrule(r){1-5}
        \cmidrule(l){6-8}  
        \textbf{n07717070} & n07717410 & n07717556 & & 
        & \textbf{n00003553} & n12057211 & n12620546 \\
        \cmidrule(r){1-5}
        \cmidrule(l){6-8}  
        \textbf{n13134947} & n12144580 & n13133613 \\
        \bottomrule[1.5pt] 
    \end{tabular}
    }
    \label{tab:super2}
\end{table*}

\section{Additional Results}
\begin{figure}
    \centering
    \includegraphics[width=\linewidth]{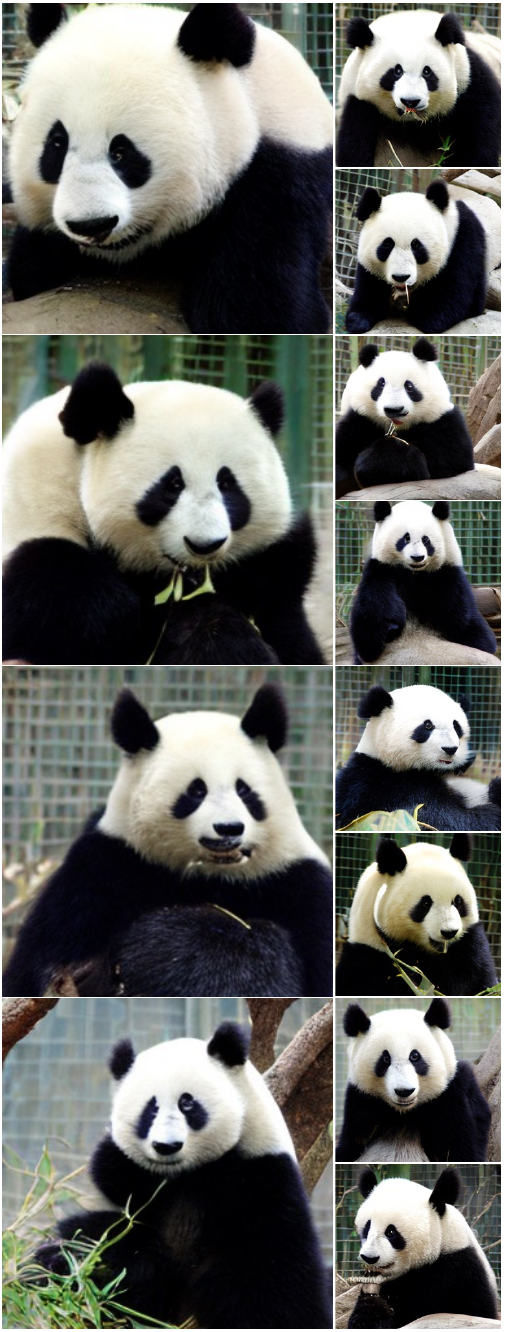}
    \vspace{-0.3in}
    \caption{Images of n02510455 generated by KIND.}
    \label{fig:more1}
\end{figure}

\begin{figure}
    \centering
    \includegraphics[width=\linewidth]{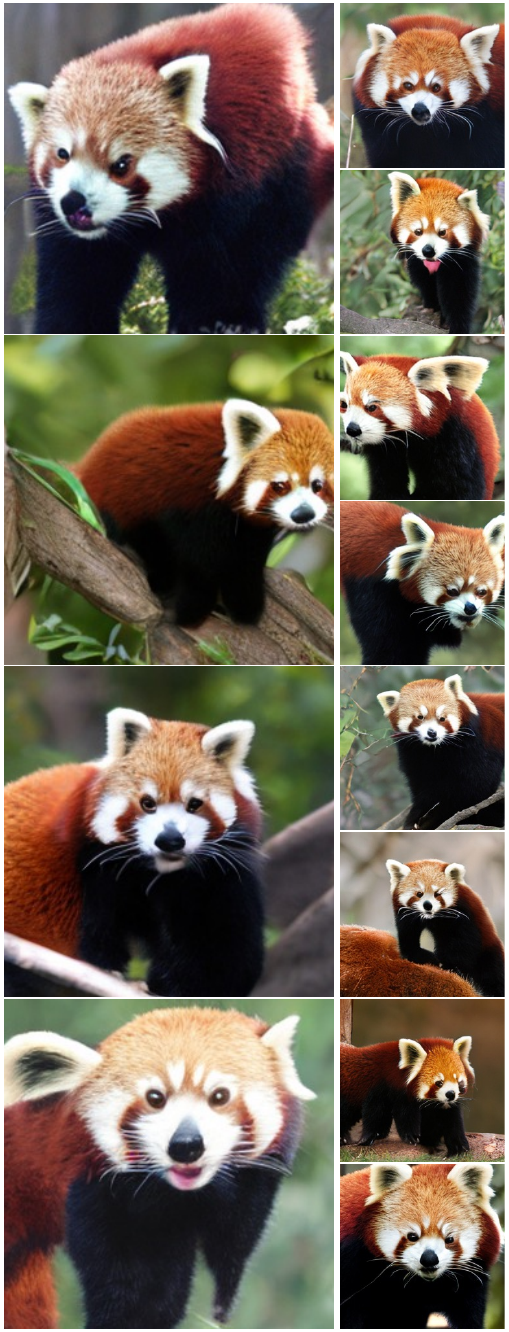}
    \vspace{-0.3in}
    \caption{Images of n02509815 generated by KIND.}
    \label{fig:more2}
\end{figure}

\begin{figure}
    \centering
    \includegraphics[width=\linewidth]{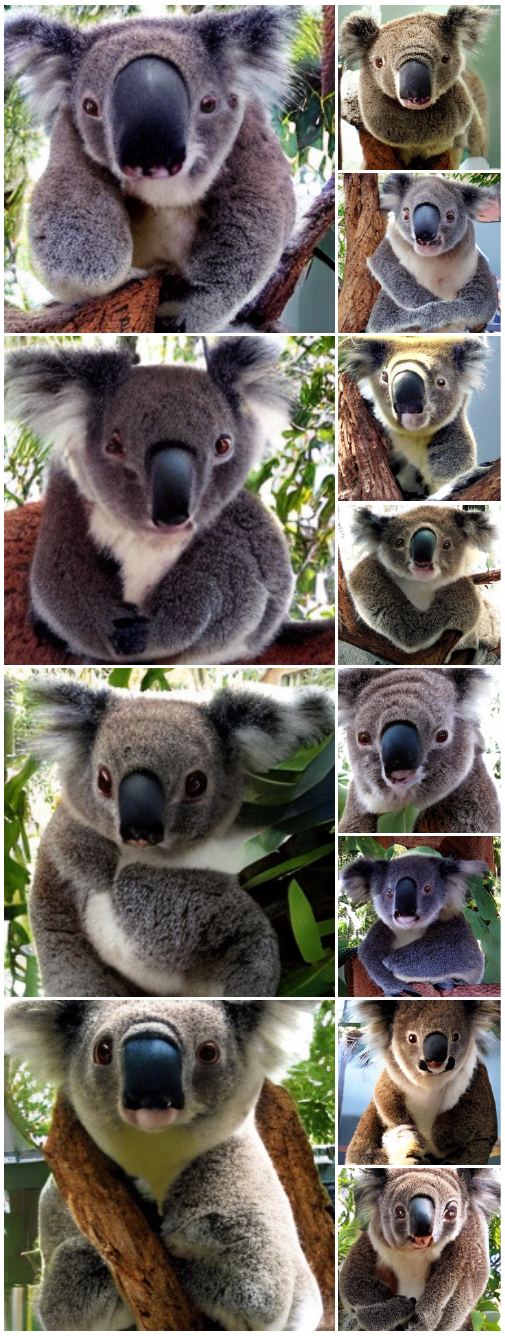}
    \vspace{-0.3in}
    \caption{Images of n01882714 generated by KIND.}
    \label{fig:more3}
\end{figure}

\begin{figure}
    \centering
    \includegraphics[width=\linewidth]{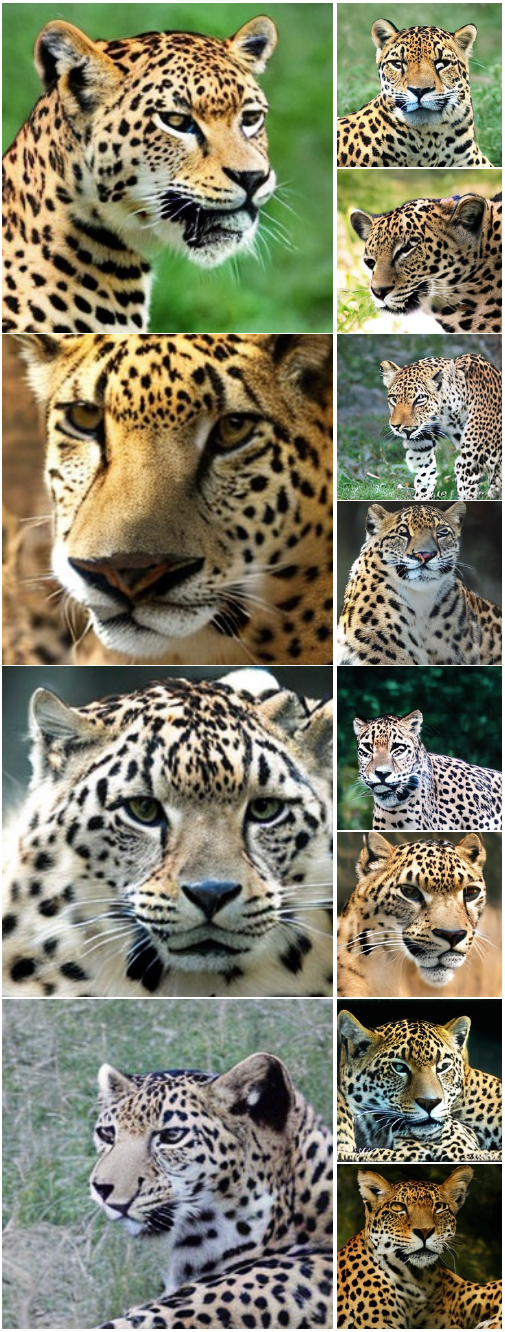}
    \vspace{-0.3in}
    \caption{Images of n02120997 generated by KIND.}
    \label{fig:more4}
\end{figure}

\begin{figure}
    \centering
    \includegraphics[width=\linewidth]{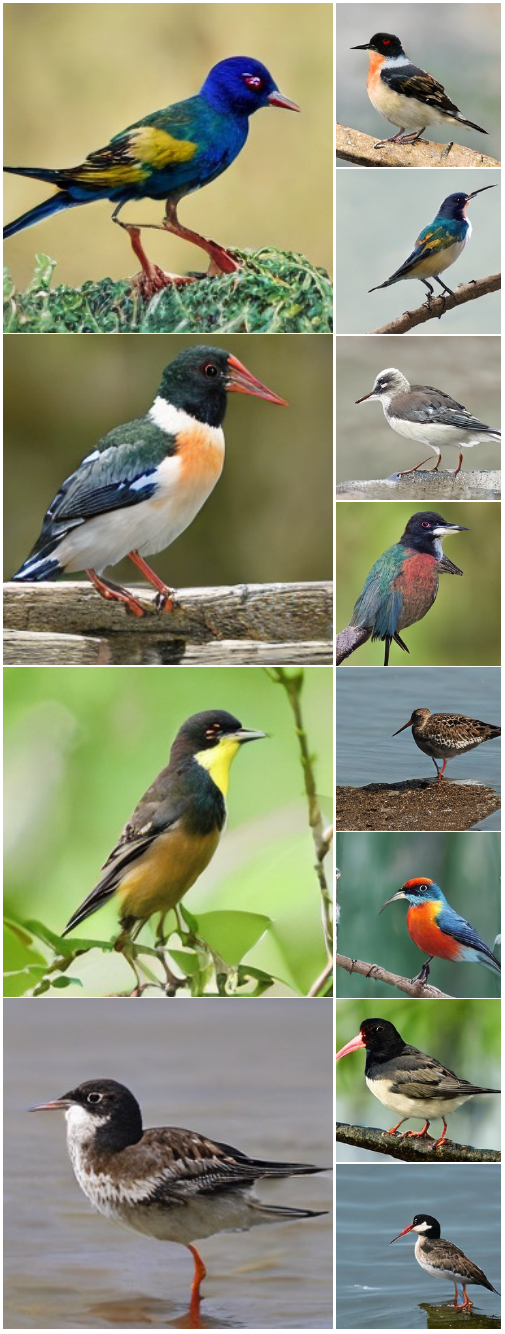}
    \vspace{-0.3in}
    \caption{Images of n01503061 generated by KIND.}
    \label{fig:more5}
\end{figure}

\begin{figure}
    \centering
    \includegraphics[width=\linewidth]{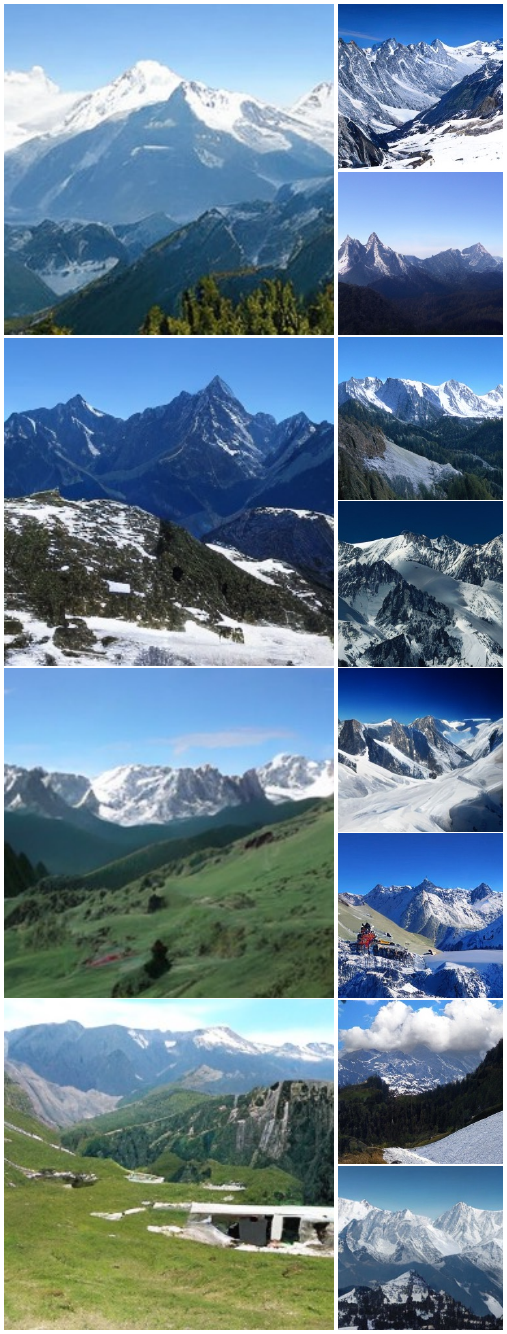}
    \vspace{-0.3in}
    \caption{Images of n09193705 generated by KIND.}
    \label{fig:more6}
\end{figure}

\begin{figure}
    \centering
    \includegraphics[width=\linewidth]{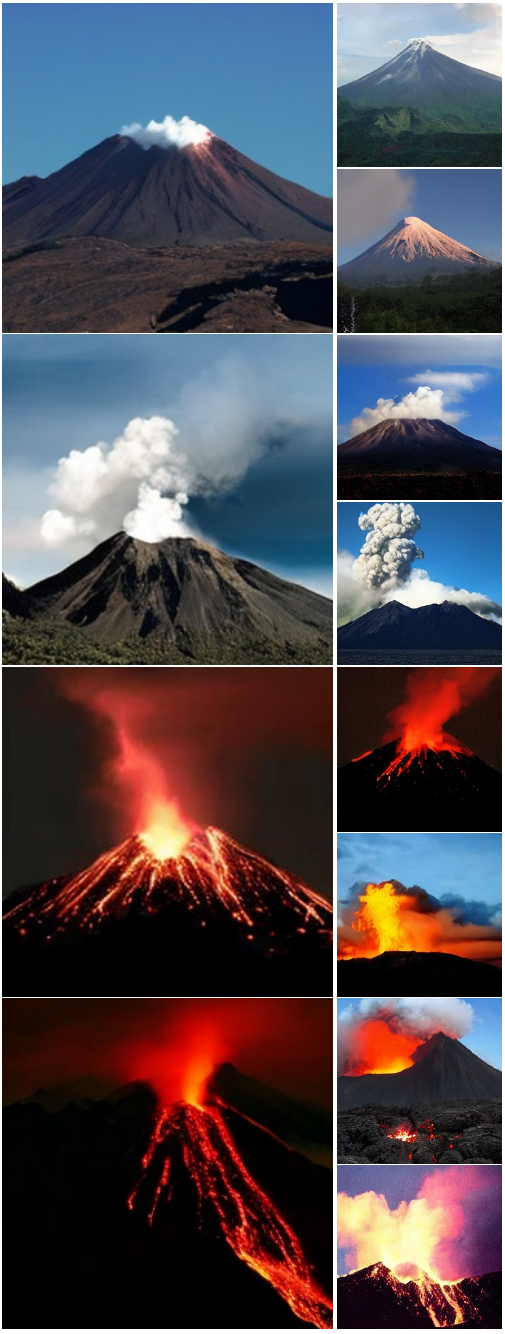}
    \vspace{-0.3in}
    \caption{Images of n09472597 generated by KIND.}
    \label{fig:more7}
\end{figure}

\begin{figure}
    \centering
    \includegraphics[width=\linewidth]{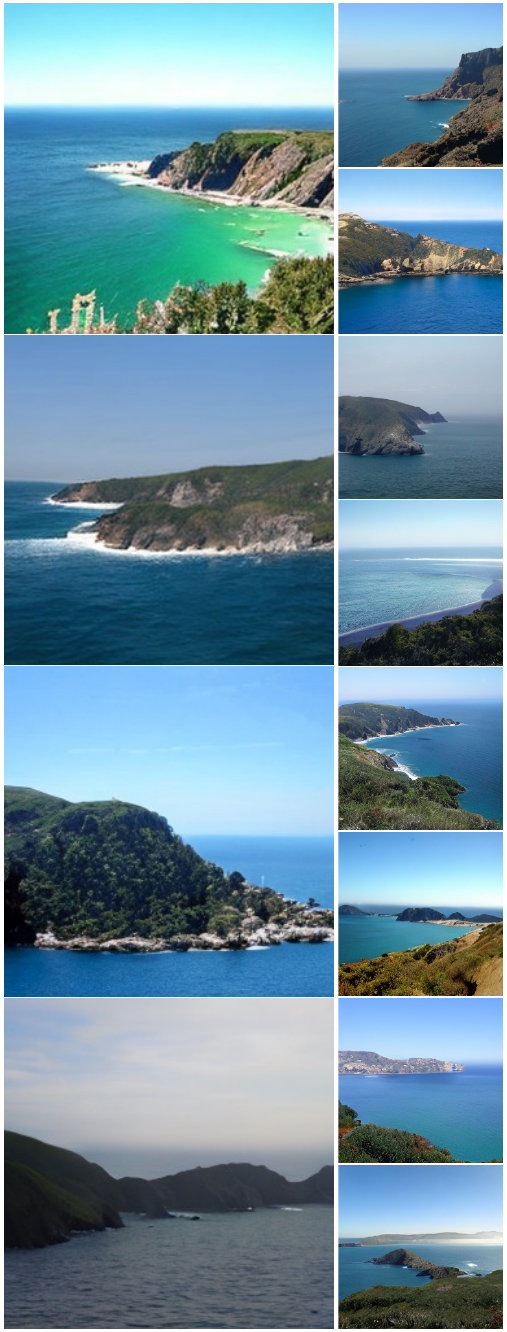}
    \vspace{-0.3in}
    \caption{Images of n09399592 generated by KIND.}
    \label{fig:more8}
\end{figure}

We provide more images of novel classes generated by our KIND which is a DiT-L/2 model composed of learngenes and tailors at 256 × 256 resolution, as shown in Figure \ref{fig:more1}-\ref{fig:more8}.
\twocolumn

\end{document}